%% file: main.tex
\PassOptionsToPackage{table}{xcolor}
\documentclass[runningheads]{llncs}

\usepackage[mobile]{eccv}

\usepackage{eccvabbrv}

\usepackage{graphicx}
\usepackage{booktabs}
\usepackage{wrapfig}

\usepackage[accsupp]{axessibility}  %

\usepackage[pagebackref,breaklinks,colorlinks,citecolor=eccvblue]{hyperref}
\usepackage{orcidlink}

\input{preamble}

\begin{document}

\title{FlowMap: High-Quality Camera Poses, Intrinsics, and Depth via Gradient Descent}

\titlerunning{FlowMap: Camera Poses, Intrinsics and Depth via Gradient Descent}

\author{Cameron Smith* \and
David Charatan* \and \\
Ayush Tewari \and
Vincent Sitzmann}

\authorrunning{C.~Smith and D.~Charatan et al.}

\institute{MIT CSAIL}

\maketitle

\input{sec/0_abstract}
\input{sec/1_intro}
\input{sec/2_related}
\input{sec/3_method}
\input{sec/4_experiments}
\input{sec/5_ablations}
\input{sec/6_conclusion}

\bibliographystyle{splncs04}
\bibliography{main}

\title{Supplemental Material}
\author{}
\institute{}
\maketitle
\vspace{-20pt}

\makeatletter
\renewcommand \thesection{S\@arabic\c@section}
\renewcommand\thetable{S\@arabic\c@table}
\renewcommand \thefigure{S\@arabic\c@figure}
\makeatother
\setcounter{section}{0}
\setcounter{figure}{0}
\setcounter{table}{0}

\input{supplement/additional_ablations}

\input{supplement/additional_results}
\input{supplement/implementation_details}
\input{supplement/experiment_details}
\input{supplement/limitations}

\end{document}

%% file: preamble.tex
\usepackage{paralist}
\usepackage{algorithm}
\usepackage{algpseudocode}
\usepackage{comment}
\usepackage{float}

\def\ints{\mathbf{K}}
\def\loss{\mathcal{L}}

\def\pixcoord{\mathbf{u}}
\def\pixcoordx{\mathbf{x}}

\DeclareMathOperator*{\argmin}{arg\,min}

\def\pose{\textbf{P}}
\def\weights{\mathcal{W}}

\def\depth{\mathbf{D}}
\def\SE{\text{SE}}
\def\surf{\mathbf{X}}

\def\flow{\mathbf{F}} %

\newcommand{\myparagraph}[1]{\vspace{-4pt}\subsubsection{#1}}

%% file: sec/0_abstract.tex
\begin{abstract}
This paper introduces FlowMap, an end-to-end differentiable method that solves for precise camera poses, camera intrinsics, and per-frame dense depth of a video sequence.
Our method performs per-video gradient-descent minimization of a simple least-squares objective that compares the optical flow induced by depth, intrinsics, and poses against correspondences obtained via off-the-shelf optical flow and point tracking.
Alongside the use of point tracks to encourage long-term geometric consistency, we introduce differentiable re-parameterizations of depth, intrinsics, and pose that are amenable to first-order optimization.
We empirically show that camera parameters and dense depth recovered by our method enable photo-realistic novel view synthesis on $360^\circ$ trajectories using Gaussian Splatting.
Our method not only far outperforms prior gradient-descent based bundle adjustment methods, but surprisingly performs on par with COLMAP, the state-of-the-art SfM method, on the downstream task of $360^\circ$ novel view synthesis---even though our method is purely gradient-descent based, fully differentiable, and presents a complete departure from conventional SfM.
\end{abstract}

%% file: sec/1_intro.tex
\section{Introduction}

\label{sec:intro}

Reconstructing a 3D scene from video is one of the most fundamental problems in vision and has been studied for over five decades.
Today, essentially all state-of-the-art approaches are built on top of Structure-from-Motion (SfM) methods like COLMAP~\cite{schonberger2016structure}. These approaches extract sparse correspondences across frames, match them, discard outliers, and then optimize the correspondences' 3D positions alongside the camera parameters by minimizing reprojection error~\cite{schonberger2016structure}.

This framework has delivered excellent results which underlie many present-day vision applications, and so it is unsurprising that SfM systems have remained largely unchanged in the age of deep learning, save for deep-learning-based correspondence matching \cite{sarlin2020superglue,lindenberger2023lightglue,sarlin2021pixloc,detone2018superpoint}.

However, conventional SfM has a major limitation: it is not differentiable with respect to its free variables (camera poses, camera intrinsics, and per-pixel depths).
This means that SfM acts as an isolated pre-processing step that cannot be embedded into end-to-end deep learning pipelines. 
A differentiable, self-supervised SfM method would enable neural networks to be trained self-supervised on internet-scale data for a broad class of multi-view geometry problems.
This would pave the way for deep-learning based 3D reconstruction and scene understanding.

\input{figures/overview/figure}
In this paper, we present FlowMap, a differentiable and surprisingly simple camera and geometry estimation method whose outputs enable photorealistic novel view synthesis. 
FlowMap directly minimizes the difference between optical flow that is induced by a camera moving through a static 3D scene and pre-computed correspondences in the form of off-the-shelf point tracks and optical flow.
Since FlowMap is end-to-end differentiable, it can naturally be embedded in any deep learning pipeline.
Its loss is minimized only via gradient descent, leading to high-quality camera poses, camera intrinsics, and per-pixel depth.
Unlike conventional SfM, which outputs sparse 3D points that are each constrained by several views, FlowMap outputs dense per-frame depth estimates.
This is a critical advantage in downstream novel view synthesis and robotics tasks.
Unlike prior attempts at gradient-based optimization of cameras and 3D geometry~\cite{lin2021barf,nerf--,bian2022nopenerf}, we do not treat depth, intrinsics, and camera poses as free variables.
Rather, we introduce differentiable feed-forward estimates of each one: depth is parameterized via a neural network, pose is parameterized as the solution to a least-squares problem involving depth and flow, and camera intrinsics are parameterized using a differentiable selection based on optical flow consistency.
In other words, FlowMap solves SfM by learning the depth network's parameters; camera poses and intrinsics are computed via analytical feed-forward modules without free parameters of their own.
We show that this uniquely enables high-quality SfM via gradient descent while making FlowMap compatible with standard deep-learning pipelines.
Unlike recent radiance-field bundle-adjustment baselines~\cite{bian2022nopenerf,lin2021barf}, FlowMap does not use differentiable volume rendering, and so it is significantly faster to run, generally reconstructing an object-centric $360^\circ$ scan in less than 10 minutes.

Through extensive ablation studies, we show that each of FlowMap's design choices is necessary.
On popular, real-world novel view synthesis datasets (Tanks \& Temples, Mip-NeRF 360, CO3D, and LLFF), we demonstrate that FlowMap enables photo-realistic novel view synthesis up to full $360^\circ$ trajectories using Gaussian Splatting~\cite{kerbl20233d}. 
Gaussian Splats obtained from FlowMap reconstructions far outperform the state-of-the-art gradient-based bundle-adjustment method, NoPe-NeRF~\cite{bian2022nopenerf}, and those obtained using the SLAM algorithm DROID-SLAM~\cite{teed2021droid}, even though both baselines require ground-truth intrinsics.
Gaussian Splats obtained from FlowMap are on par with those obtained from COLMAP~\cite{schonberger2016structure}, even though FlowMap only leverages gradient descent, is fully differentiable, and represents a complete departure from conventional SfM techniques.

%% file: figures/overview/figure.tex
\begin{figure*}[t!]
    \centering
    \includegraphics[width=\linewidth]{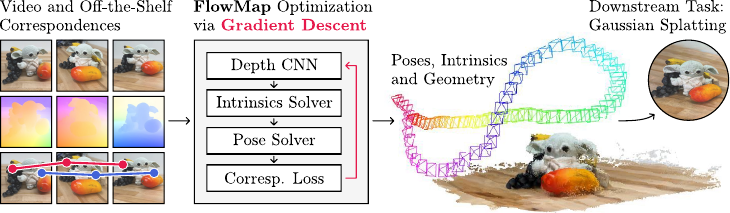}
    \caption{
    We present FlowMap, an end-to-end differentiable method that recovers poses, intrinsics, and depth maps of an input video.
    FlowMap is supervised only with off-the-shelf optical flow and point track correspondences, and optimized per-scene with gradient descent.
    Gaussian Splats obtained from FlowMap's reconstructions regularly match or exceed those obtained from COLMAP in quality.
    } 
    \label{fig:overview}
    \vspace{-20pt}
\end{figure*}

%% file: sec/2_related.tex
\section{Related Work}
\label{sec:related}

\myparagraph{Conventional Structure-from-Motion (SfM) and SLAM.}
Modern SfM methods perform offline optimization using a multi-stage process of descriptor extraction, correspondence estimation, and subsequent incremental bundle adjustment.
In bundle adjustment, corresponding 2D pixels are coalesced into single 3D points, and estimated camera parameters are optimized alongside these points' 3D positions to minimize 3D-to-2D reprojection error.
COLMAP~\cite{schonberger2016structure} is the de-facto standard for accurate, offline camera parameter estimation.
Meanwhile, simultaneous localization and mapping (SLAM) usually refers to real-time, online methods. These generally assume that the camera's intrinsic parameters are known.
Similar to SfM, SLAM usually relies on minimizing reprojection error~\cite{mur2015orb,mur2017orb,campos_orb3,rosinol2020kimera}, but some methods investigate direct minimization of a photometric error~\cite{engel2014lsd,engel2017direct}.
While deep learning has not fundamentally transformed SfM and SLAM, it has been leveraged for correspondence prediction \cite{choy2016universal,mishchuk2017working,luo2018geodesc,ono2018lf}, either via graph neural networks~\cite{sarlin2020superglue} or via particle tracking~\cite{zhao2022particlesfm, harley2022particle, doersch2023tapir}.

FlowMap is a departure from conventional SfM and SLAM techniques.
While we rely on correspondence from optical flow and particle tracking, we do not coalesce sets of 2D correspondences into single 3D points.
Instead, we use per-frame depth estimates as our geometry representation.
Additionally, rather than relying on conventional correspondence matching and RANSAC filtering, we leverage neural point tracking~\cite{karaev2023cotracker} and optical flow estimators~\cite{raft} to establish correspondence, jointly enabling dense geometry reconstruction without a seperate multi-view stereo stage.
Finally, FlowMap is end-to-end differentiable and introduces feed-forward estimators of depth, poses, and intrinsics, making it compatible with other learned methods.

\myparagraph{Deep-Learning Based SfM.}
Prior work has attempted to embed the full SLAM pipeline into a deep learning framework~\cite{tang2018ba,czarnowski2020deepfactors,zhou2018deeptam,clark2018learning,ummenhofer2017demon,liu2019neural,teed2018deepv2d,wang2021tartanvo,bloesch2018codeslam}, usually by training black-box neural networks to directly output camera poses. 
However, these methods are constrained to short videos of 5 to 10 frames and are not competitive with conventional SLAM and SfM for real-world 3D reconstruction.
Bowen et al.\cite{bowen2022dimensions} elegantly leverage optical flow supervision for self-supervised monocular depth prediction.
More recently, DROID-SLAM~\cite{teed2021droid} has yielded high-quality camera poses and depth.
However, it requires known intrinsics, is trained fully supervised with ground-truth camera poses, and fails to approach COLMAP on in-the-wild performance and robustness.
Concurrent work to FlowMap explores an end-to-differentiable, point-tracking-based SfM framework~\cite{wang2023visual}.
Key differences are that their method is fully supervised with camera poses, point clouds, and intrinsics; requires large-scale, multi-stage training; solves only for sparse depth; and is built around the philosophy of making each part of the conventional SfM pipeline differentiable.
Our method is a complete departure from the conventional SfM pipeline---it does not require a training set of known intrinsics, ground-truth poses, or 3D points, and it provides quality gradients for dense depth, poses, and intrinsics.
Critically, FlowMap is the first gradient-descent based method to rival the performance of conventional SfM on the novel view synthesis task.
Zhang et al. \cite{zhang2022casual} and Kopf ef al. \cite{kopf2021robust} demonstrate gradient-descent based optimization of camera parameters with a similar flow-based reprojection supervision, with a focus on dynamic scenes. 
However, these methods optimize camera parameters as free variables and depend on pre-trained monocular depth estimators. 
In constrast, our feed-forward parameterization uniquely enables gradients for large-scale training and we demonstrate that our gradients can be used to \textit{train} a depth estimator.

\myparagraph{Novel View Synthesis via Differentiable Rendering.}
Advances in differentiable rendering have enabled photo-realistic novel view synthesis and fine-grained geometry reconstruction using camera poses and intrinsics obtained via SfM~\cite{mildenhall2020nerf, li2023neuralangelo, mueller2022instant, sitzmann2019deepvoxels, sitzmann2019scene, niemeyer2020differentiable}. 
3D Gaussian Splatting~\cite{kerbl20233d} goes further, directly leveraging the 3D points provided by SfM as an initialization.
It follows previous methods like~\cite{dsnerf}, which used 3D geometry from depth to supervise neural radiance field (NeRF) reconstructions.
We show that when initializing Gaussian Splatting with poses, intrinsics, and 3D points from FlowMap, we generally perform on par with conventional SfM and sometimes even outperform it.

\myparagraph{Camera Pose Optimization via Differentiable Rendering.}
A recent line of work in bundle-adjusting radiance fields \cite{lin2021barf,bian2022nopenerf,yugay2023gaussian,keetha2023splatam,jeong2021self,fu2023cbarf,wu2023scanerf,fu2023colmapfree,yen2021inerf,xia2022sinerf,wang2021nerf,chng2022garf,cheng2023lu} attempts to jointly optimize unknown camera poses and radiance fields.
Several of these methods ~\cite{jeong2021self, yan2023cf, 10377535} additionally solve for camera intrinsic parameters.
However, these methods only succeed when given forward-facing scenes or roughly correct pose initializations.
More recent work incorporates optical flow and monocular depth priors \cite{bian2022nopenerf,meuleman2023localrf,liu2023robust} but requires known intrinsics and only works robustly on forward-facing scenes.
Concurrent work \cite{fu2023colmapfree} accelerates optimization compared to earlier NeRF-based approaches. Unlike ours, this approach requires known intrinsics and a pre-trained monocular depth estimator, and minimizes a volume-rendering-based photometric loss instead of the proposed correspondence-based approach.
Further concurrent work proposes real-time SLAM via gradient descent on 3D Gaussians~\cite{Matsuki:Murai:etal:CVPR2024}, but requires known intrinsics and does not show robustness on a variety of real-world scenes.
In contrast, our method is robust and easily succeeds on object-centric scenes where the camera trajectory covers a full $360^{\circ}$ of rotation, yielding photo-realistic novel view synthesis when combined with Gaussian Splatting.

\myparagraph{Learning Priors over Optimization of NeRF and Poses.}
Our method is inspired by recent methods which learn priors over pose estimation and 3D radiance fields~\cite{Chen_2023_CVPR, lai2021video, smith2023flowcam, fu2022mononerf}. 
However, these approaches require known camera intrinsics, are constrained to scenes with simple motion, and do not approach the accuracy of conventional SfM. 
Like our method, FlowCam~\cite{smith2023flowcam} uses a pose-induced flow loss and a least-squares solver for camera pose.
However, our method has several key differences: we estimate camera intrinsics, enabling optimization on any raw video; we replace 3D rendering with a simple depth estimator, which reduces training costs and allows us to reuse pre-trained depth estimators; and we introduce point tracks for supervision to improve global consistency and reduce drift.
FlowCam did not approach conventional SfM's accuracy on real sequences.
We demonstrate that optimizing the pose-induced flow objective on a single scene, akin to a test-time optimization, yields pose and geometry estimates which, for the first time, approach COLMAP on full $360^\circ$ sequences.

%% file: sec/3_method.tex
\input{figures/flowchart/figure}
\section{Supervision via Camera-Induced Scene Flow}
\label{sec:loss}
Given a video sequence, our goal is to supervise per-frame estimates of depth, intrinsics, and pose using known correspondences.
Our method hinges upon the fact that a camera moving through a static scene induces optical flow in image space.
Such optical flow can be computed differentiably from any two images' estimated depths, intrinsics, and relative pose to yield a set of implied pixel-wise correspondences.
These correspondences can then by compared to their known counterparts to yield supervision on the underlying estimates.

Consider a 2D pixel at coordinate $\pixcoord_i \in \mathbb{R}^2$ in frame $i$ of the video sequence. 
Using frame $i$'s estimated depth $\depth_i$ and intrinsics $\ints_i$, we can compute the pixel's 3D location $\pixcoordx_i \in \mathbb{R}^3$.
Then, using the estimated relative pose $\pose_{ij}$ between frames $i$ and $j$, we can transform this location into frame $j$'s camera space.
Finally, we can project the resulting point $\pose_{ij} \pixcoordx_i$ onto frame $j$'s image plane to yield an implied correspondence $\hat\pixcoord_{ij}$.
This correspondence can be compared to the known correspondence $\pixcoord_{ij}$ to yield a loss $\loss$, as illustrated in Fig.~\ref{fig:loss}.
\begin{equation}
    \loss = \| \hat\pixcoord_{ij} - \pixcoord_{ij} \|
    \label{eq:loss}
\end{equation}
\vspace{-25pt}
\myparagraph{Supervision via Dense Optical Flow and Sparse Point Tracks.}
Our known correspondences are derived from two sources: dense optical flow between adjacent frames and sparse point tracks which span longer windows.
Frame-to-frame optical flow ensures that depth is densely supervised, while point tracks minimize drift over time.
We compute correspondences from optical flow $\flow_{ij}$ via $\pixcoord_{ij} = \pixcoord_i + \flow_{ij}[\pixcoord_i]$.
Meanwhile, given a query point $\pixcoord_i$, an off-the-shelf point tracker directly provides a correspondence $\pixcoord_{ij}$ for any frame $j$ where one exists.

\input{figures/loss/figure}
\myparagraph{Baseline: Pose, Depth and Intrinsics as Free Variables.}
Assuming one uses standard gradient descent optimization, one must decide how to parameterize the estimated depths, intrinsics, and poses.
The simplest choice is to parameterize them as free variables, i.e., to define learnable per-camera intrinsics and extrinsics alongside per-pixel depths.
However, this approach empirically fails to converge to good poses and geometry, as shown in Sec.~\ref{sec:ablations}.

\section{Parameterizing Depth, Pose, and Camera Intrinsics}
\label{sec:reparams}
In this section, we present FlowMap's feed-forward re-parameterization of depth, pose, and camera intrinsics, which uniquely enables high-quality results when using gradient descent. 
Later, in Sec.~\ref{sec:ablations}, we ablate these parameterizations to demonstrate that they lead to dramatic improvements in accuracy.

\myparagraph{Depth Network.} 
If each pixel's depth were optimized freely, two identical or very similar image patches could map to entirely different depths.
We instead parameterize depth as a neural network that maps an RGB frame to the corresponding per-pixel depth. 
This ensures that similar patches have similar depths, allowing FlowMap to integrate geometry cues across frames: if a patch receives a depth gradient from one frame, the weights of the depth network are updated, and hence the depths of all similar video frame patches are also updated.
As a result, FlowMap can provide high-quality depths even for patches which are poorly constrained due to errors in the input flows and point tracks, imperceptibly small motion, or degenerate (rotation-only) motion.

\myparagraph{Pose as a Function of Depth, Intrinsics and Optical Flow.}
\input{figures/procrustes/figure}
Suppose that for two consecutive frames, optical flow, per-pixel depths, and camera intrinsics are known.
In this case, the relative pose between these frames can be computed differentiably in closed form.
Following the approach proposed in FlowCam~\cite{smith2023flowcam}, we solve for the relative pose that best aligns each consecutive pair of un-projected depth maps.
We then compose the resulting relative poses to produce absolute poses in a common coordinate system.

More formally, we cast depth map alignment as an orthogonal Procrustes problem, allowing us to draw upon this problem's differentiable, closed-form solution~\cite{choy2020deep}.
We begin by unprojecting the depth maps $\depth_i$ and $\depth_j$ using their respective intrinsics $\ints_i$ and $\ints_j$ to generate two point clouds $\surf_i$ and $\surf_j$.
Next, because the Procrustes formulation requires correspondence between points, we use the known optical flow between frames $i$ and $j$ to match points in $\surf_i$ and $\surf_j$.
This yields $\surf_i^\leftrightarrow$ and $\surf_j^\leftrightarrow$, two filtered point clouds for which a one-to-one correspondence exists.
The Procrustes formulation seeks the rigid transformation that minimizes the total distance between the matched points:
\begin{align}
    \pose_{ij} = \underset{\pose \in \SE(3)}{\argmin} \| \weights^{1/2} (\surf_j^\leftrightarrow - \pose \surf_i^\leftrightarrow )\|_2^2
    \label{eq:procrustes}
\end{align}
The diagonal matrix $\weights$ contains correspondence weights that can down-weight correspondences that are faulty due to occlusion or imprecise flow.
This weighted least-squares problem can be solved in closed form via a single singular value decomposition~\cite{choy2020deep,smith2023flowcam} which is both cheap and fully differentiable.
We further follow FlowCam~\cite{smith2023flowcam} and predict these weights by concatenating corresponding per-pixel features and feeding them into a small MLP. This MLP's parameters are the only other free variables of our model.
For an overview of the depth map alignment process, see Fig.~\ref{fig:procrustes}.

\myparagraph{Camera Focal Length as a Function of Depth and Optical Flow.}

We solve for camera intrinsics by considering a set of reasonable candidates $\ints_k$, then softly selecting among them.
For each candidate, we use our pose solver Eq.~\ref{eq:procrustes} to compute a corresponding set of poses, then use the camera-induced flow loss Eq.~\ref{eq:loss} to compute the loss $\loss_k$ implied by $\ints_k$ and these poses.
Finally, we compute the resulting intrinsics $\ints$ via a softmin-weighted sum of the candidates:
\begin{align}
    \ints = \sum_k w_k \ints_k && w_k = \frac{\exp(-\loss_k)}{\sum_l \exp(-\loss_l)}
\end{align}
To make this approach computationally efficient, we make several simplifying assumptions.
First, we assume that the intrinsics can be represented via a single $\ints$ that is shared across frames.
Second, we assume that $\ints$ can be modeled via a single focal length with a principal point fixed at the image center.
Finally, we only compute the soft selection losses on the first two frames of the sequence.

\myparagraph{Depth as the Only Free Variable in SfM.} FlowMap offers a surprising insight: Given correspondence, SfM can be formulated as solving for per-frame depth maps.
FlowMap yields poses and intrinsics in a parameter-free, differentiable forward pass when given correspondences and depths.
This means that better initializations of FlowMap's depth estimator (e.g., from pre-training) will yield more accurate camera parameters (see Fig.~\ref{fig:prior}).

\input{figures/splat_comparison/figure}

\section{Implementation and Optimization Details}
\label{sec:optimization}
FlowMap is optimized on each specific scene, achieving convergence between 500 and 5,000 steps using the Adam~\cite{kingma2014adam} optimizer.
Though per-scene optimization is key to achieving high accuracy, we find that exploiting FlowMap's feed-forward nature for pre-training yields an initialization that leads to improved convergence and accuracy, as shown in Fig.~\ref{fig:prior}.
During per-scene optimization, we use RAFT~\cite{raft} and CoTracker V1~\cite{karaev2023cotracker} to compute the optical flow and point tracks that FlowMap uses as input.
During pre-training, in order to minimize the time spent computing correspondences, we do not use point tracks and use GMFlow~\cite{xu2022gmflow} to compute optical flow instead of RAFT.

\myparagraph{Focal Length Regression.}

While our soft selection approach robustly yields near-correct focal lengths, its performance is slightly worse compared to well-initialized direct regression.
We therefore switch to focal length regression after 1,000 steps, using our softly selected focal length as initialization.

\myparagraph{Memory and Time Requirements.}

FlowMap's complexity in time and memory is linear with the number of input video frames.
During each optimization step, FlowMap recomputes depth for each frame, then derives poses and intrinsics from these depths to generate gradients.
In practice, FlowMap optimization for a 150-frame video takes about 20 minutes, with a peak memory usage of about 36 GB.
Precomputing point tracks and optical flow takes approximately 2 minutes.
Note that FlowMap's runtime could be reduced by early stopping, and its memory usage could be reduced by performing backpropagation on video subsets during each step, but we leave these optimizations to future work.

\myparagraph{Sequence Length and Drift.} 
Since adjacent frames in typical 30 FPS videos usually contain redundant information, we run FlowMap on subsampled videos.
We perform subsampling by computing optical flow on the whole video, then selecting frames so as to distribute the overall optical flow between them as evenly as possible.
With this strategy, we find that an object-centric, full $360^\circ$ trajectory as is common in novel view synthesis papers is covered by about $90$ frames.
We note that FlowMap does not have a loop closure mechanism.
Rather, point tracks provide long-range correspondences that prevent the accumulation of drift in long sequences.

%% file: figures/flowchart/figure.tex
\begin{figure*}[t!]
    \centering
    \includegraphics[width=\linewidth,]{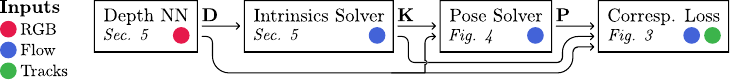}
    \caption{\textbf{A FlowMap Forward Pass.}
    Given RGB frames (red), optical flow (blue) and point tracks (green), FlowMap computes dense depth $\depth$, camera poses $\pose$, and intrinsics $\ints$ in each forward pass. 
    We obtain depth via a CNN (\cref{sec:reparams}) and implement differentiable, feed-forward solvers for intrinsics and poses (\cref{sec:reparams}, Fig.\ref{fig:procrustes}).
    Colored dots indicate which block receives which inputs.
    FlowMap's only free parameters are the weights of a depth NN and a small correspondence confidence MLP.
    These parameters are optimized for each video separately by minimizing a camera-induced flow loss (Fig.~\ref{fig:loss}) via gradient descent, though fully feed-forward operation is possible.
    }
    \label{fig:flowchart}
    \vspace{-10pt}
\end{figure*}

%% file: figures/loss/figure.tex
\begin{figure*}[t]
    \centering
    \includegraphics[width=\linewidth,]{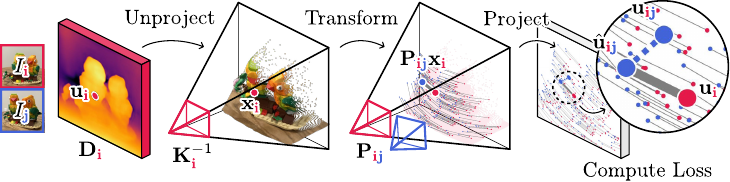}
    \caption{\textbf{Camera-Induced Flow Loss.} To use a known correspondence $\pixcoord_{ij}$ to compute a loss $\loss$, we unproject $\pixcoord_i$ using the corresponding depth map $\depth_i$ and camera intrinsics $\textbf{K}_i$, transform the resulting point $\pixcoordx_i$ via the relative pose $\mathbf{P}_{ij}$, reproject the transformed point to yield $\hat{\pixcoord}_{ij}$, and finally compute $\loss = \|\hat{\pixcoord}_{ij} - \pixcoord_{ij}\|$.}
    \label{fig:loss}
    \vspace{-10pt}
\end{figure*}

%% file: figures/procrustes/figure.tex
\begin{figure*}[t]
    \centering
    \includegraphics[width=\linewidth,]{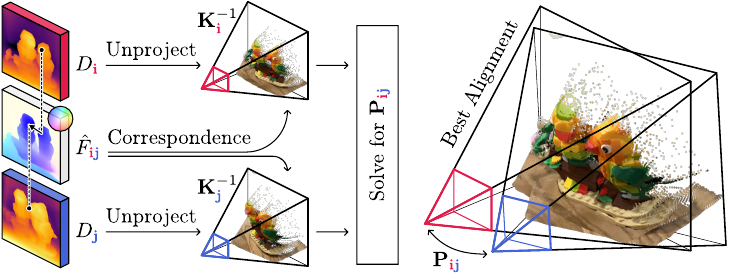}
    \vspace{-12pt}
    \caption{We solve for the relative poses between consecutive frames using their depth maps, camera intrinsics, and optical flow. To do so, we first unproject their depth maps, then solve for the pose that best aligns the resulting point clouds.}
    \label{fig:procrustes}
    \vspace{-15pt}
\end{figure*}

%% file: figures/splat_comparison/figure.tex
\begin{figure*}[t!]
    \centering
    \includegraphics[width=\linewidth]{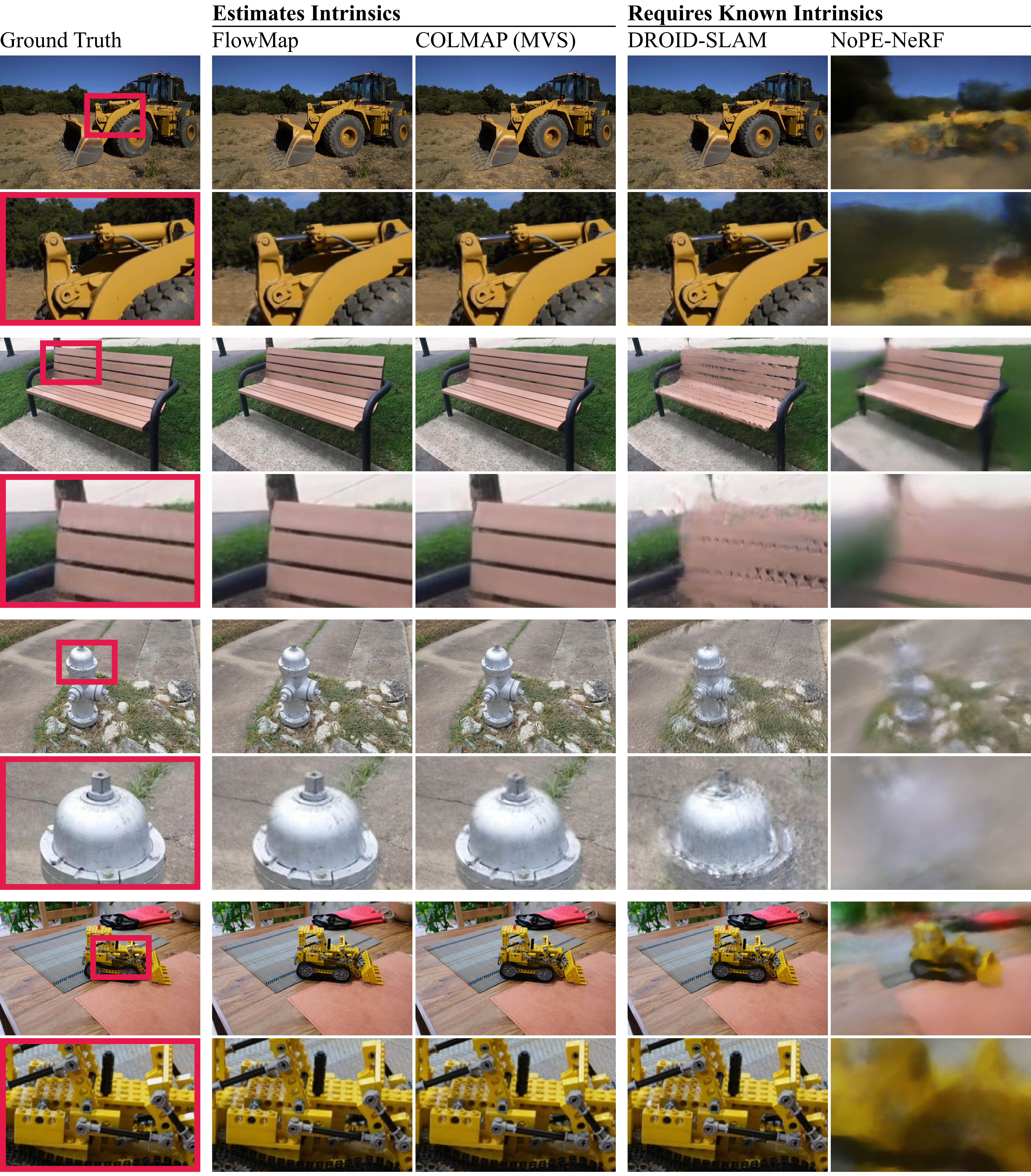}
    \vspace{-15pt}
    \caption{\textbf{View Synthesis Results.} FlowMap's camera parameters and geometry produce near-photorealistic 3D Gaussian Splatting results on par with COLMAP's.}
    \label{fig:splat_comparison}
    \vspace{-5pt}
\end{figure*}

%% file: sec/4_experiments.tex
\input{tables/table_main_comparison}
\input{figures/trajectories/figure}
\input{figures/point_clouds}

\section{Results}
\label{sec:exp}

We benchmark FlowMap via the downstream task of 3D Gaussian reconstruction~\cite{kerbl20233d}.
This allows us to measure the quality of the camera parameters and geometry (depth maps) it outputs \emph{without having access to ground-truth scene geometry and camera parameters}.

\myparagraph{Baselines.}
We benchmark FlowMap against several baselines.
First, we evaluate against COLMAP~\cite{schonberger2016structure}, the state-of-the-art structure-from-motion (SfM) method.
Given a collection of images, COLMAP outputs per-image camera poses and intrinsics alongside a sparse 3D point cloud of the underlying scene.
3D Gaussian Splatting, which was designed around COLMAP's SfM outputs, is initialized using this point cloud.
Second, we evaluate against COLMAP multi-view stereo (MVS), which enhances COLMAP's output with a much denser 3D point cloud.
When initialized using this denser point cloud, 3D Gaussian Splatting produces slightly better results.
However, note that COLMAP MVS is rarely used in practice because it can be prohibitively time-consuming to run.
Third, we evaluate against DROID-SLAM, a neural SLAM system trained on a synthetic dataset of posed video trajectories.
Finally, we evaluate against NoPE-NeRF, an method that jointly optimizes a neural radiance field and unknown camera poses.
Note that unlike FlowMap and COLMAP, both DROID-SLAM and NoPE-NeRF require camera intrinsics as input.

\myparagraph{Datasets.}

We analyze FlowMap on four standard novel view synthesis datasets: MipNeRF-360~\cite{barron2021mipnerf}, Tanks \& Temples~\cite{Knapitsch2017tanks}, LLFF~\cite{mildenhall2019local}, and CO3D~\cite{reizenstein2021common}.
Because FlowMap runs on video sequences, we restrict these datasets to just the video-like sequences they provide.

\myparagraph{Methodology.}

We run FlowMap and the baselines using images that have been rescaled to a resolution of about 700,000 pixels.
We then optimize 3D Gaussian scenes for all methods except NoPE-NeRF, since it provides its own NeRF renderings.
We use 90\% of the available views for training and 10\% for testing.
During 3D Gaussian fitting, we follow the common~\cite{nerfstudio} practice of fine-tuning the initial camera poses and intrinsics.
Such refinement is beneficial because the camera poses produced by SfM algorithms like COLMAP are generally not pixel-perfect~\cite{park2023camp,lin2021barf}.
We use the 3D points provided by COLMAP, DROID-SLAM, and FlowMap as input to 3D Gaussian Splatting.
For FlowMap, we combine the output depth maps, poses, and intrinsics to yield one point per depth map pixel.

\subsection{Novel View Synthesis Results}

Tab.~\ref{tab:recon} reports rendering quality metrics (PSNR, SSIM, and LPIPS) on the held-out test views, and Fig.~\ref{fig:splat_comparison} shows qualitative results.
Qualitatively, FlowMap facilitates high-quality 3D reconstructions with sharp details.
Quantitatively, FlowMap performs slightly better than COLMAP SfM and significantly outperforms DROID-SLAM and NoPE-NeRF.
Only COLMAP MVS slightly exceeds FlowMap in terms of reconstruction quality.
As noted previously, COLMAP MVS is rarely used for 3D Gaussian Splatting, since it is very time-consuming to run on high-resolution images.

\subsection{Camera Parameter Estimation Results}

Since the datasets we use do not provide ground-truth camera parameters, they cannot be used to directly evaluate camera parameter estimates.
Instead, Tab.~\ref{tab:recon} reports the average trajectory error (ATE) of FlowMap, DROID-SLAM, and NoPe-NeRF with respect to COLMAP.
Since COLMAP's poses are not perfect~\cite{park2023camp}, this comparison is not to be understood as a benchmark, but rather as an indication of how close these methods' outputs are to COLMAP's state-of-the-art estimates.
We find that DROID-SLAM and FlowMap both recover poses that are close to COLMAP's, while NoPE-NeRF's estimated poses are far off.
When computing ATEs, we normalize all trajectories such that $\text{tr}(XX^T) = 1$, where $X$ is an $n$-by-3 matrix of camera positions.

Fig.~\ref{fig:trajectories} plots trajectories recovered by FlowMap against those recovered by COLMAP, showing that they are often nearly identical.
Fig.~\ref{fig:point_clouds} shows point clouds derived from FlowMap's estimated depth maps and camera parameters, illustrating that FlowMap recovers well-aligned scene geometry.

\subsection{Large-Scale Robustness Study}

\input{figures/colmap_comparison}

We study FlowMap's robustness by using it to estimate camera poses for 420 CO3D scenes from 10 categories.
We compare these trajectories to CO3D's pose annotations, which were computed using COLMAP.
Since the quality of CO3D's ground-truth trajectories varies between categories, we focus on categories that have been used to train novel view synthesis models~\cite{tewari2023diffusion,chan2023generative,wewer24latentsplat}, where pose accuracy is expected to be higher.
We find that FlowMap's mean ATE (0.0056) is lower than DROID-SLAM's (0.0082) and similar to the mean ATE obtained by re-running COLMAP and comparing the results to the provided poses (0.0038).
This demonstrates that FlowMap consistently estimates poses which are close to COLMAP's.
We note that COLMAP failed to estimate poses for 36 scenes, possibly because we ran it at a sparser frame rate to be consistent with our method or because the original annotations were generated using different COLMAP settings; we exclude COLMAP's failures from the above mean ATE.
See Fig.~\ref{fig:colmap_study} for distributions of ATE values with respect to CO3D's provided camera poses.
\vspace{-10pt}

%% file: tables/table_main_comparison.tex
\setlength{\tabcolsep}{8pt}
\begin{table*}[t]

\newcommand{\first}{\cellcolor{red!40}}
\newcommand{\second}{\cellcolor{orange!40}}
\newcommand{\third}{\cellcolor{yellow!40}}
\setlength{\tabcolsep}{4pt}

\centering
\resizebox{\textwidth}{!}{

\begin{tabular}{l|rrrrr|rrrrr}
\toprule
\multicolumn{1}{c|}{} & \multicolumn{5}{|c|}{MipNeRF 360 (3 scenes)} & \multicolumn{5}{|c}{LLFF (7 scenes)} \\
\midrule
Method       & PSNR $\uparrow$ & SSIM $\uparrow$ & LPIPS $\downarrow$ & Time (min.) $\downarrow$ & ATE     & PSNR $\uparrow$ & SSIM $\uparrow$ & LPIPS $\downarrow$ & Time (min.) $\downarrow$ & ATE     \\
\midrule
FlowMap      &   \third{29.84} &   \third{0.916} &      \third{0.073} &             \third{19.8} & 0.00055 &  \second{27.23} &   \third{0.849} &     \second{0.079} &              \third{7.5} & 0.00209 \\
COLMAP       &  \second{29.95} &  \second{0.928} &              0.074 &             \second{4.8} &     N/A &           25.73 &  \second{0.851} &              0.098 &             \second{1.1} &     N/A \\
COLMAP (MVS) &   \first{31.03} &   \first{0.938} &      \first{0.060} &                     42.5 &     N/A &   \first{27.99} &   \first{0.867} &      \first{0.072} &                     13.4 &     N/A \\
DROID-SLAM*  &           29.83 &           0.913 &     \second{0.066} &              \first{0.6} & 0.00017 &   \third{26.21} &           0.818 &      \third{0.094} &              \first{0.3} & 0.00074 \\
NoPE-NeRF*   &           13.60 &           0.377 &              0.750 &                   1913.1 & 0.04429 &           17.35 &           0.490 &              0.591 &                   1804.0 & 0.03920 \\
\midrule
\multicolumn{1}{c|}{} & \multicolumn{5}{|c|}{Tanks \& Temples (14 scenes)} & \multicolumn{5}{|c}{CO3D (2 scenes)} \\
\midrule
Method       & PSNR $\uparrow$ & SSIM $\uparrow$ & LPIPS $\downarrow$ & Time (min.) $\downarrow$ & ATE     & PSNR $\uparrow$ & SSIM $\uparrow$ & LPIPS $\downarrow$ & Time (min.) $\downarrow$ & ATE     \\
\midrule
FlowMap      &  \second{27.00} &  \second{0.854} &     \second{0.101} &             \third{22.3} & 0.00124 &   \first{31.11} &   \first{0.896} &      \first{0.064} &             \third{22.1} & 0.01589 \\
COLMAP       &   \third{26.74} &   \third{0.848} &      \third{0.130} &             \second{5.5} &     N/A &           25.17 &           0.750 &              0.190 &            \second{12.6} &     N/A \\
COLMAP (MVS) &   \first{27.43} &   \first{0.863} &      \first{0.097} &                     51.4 &     N/A &   \third{25.35} &   \third{0.762} &      \third{0.175} &                     52.0 &     N/A \\
DROID-SLAM*  &           25.70 &           0.824 &              0.133 &              \first{0.8} & 0.00122 &  \second{25.97} &  \second{0.790} &     \second{0.139} &              \first{0.8} & 0.01728 \\
NoPE-NeRF*   &           13.38 &           0.449 &              0.706 &                   2432.9 & 0.03709 &           14.97 &           0.400 &              0.770 &                   2604.9 & 0.03648 \\
\bottomrule
\end{tabular}
}

\vspace{5pt}
\caption{Camera parameter and geometry intializations from FlowMap produce 3D Gaussian reconstruction results that far outperform prior gradient-based baselines and are generally on par with those produced by COLMAP. Methods marked with an asterisk require ground-truth intrinsics. We report ATE with respect to COLMAP's pose estimates for reference, since no ground-truth trajectories exist for common view synthesis datasets. We exclude scenes where COLMAP or FlowMap fail entirely; each fails on 4 scenes. See the supplementary document for more details.}
\label{tab:recon}
\vspace{-11pt}
\end{table*}

%% file: figures/trajectories/figure.tex
\begin{figure*}[t!]
    \centering
   \includegraphics[width=\linewidth]{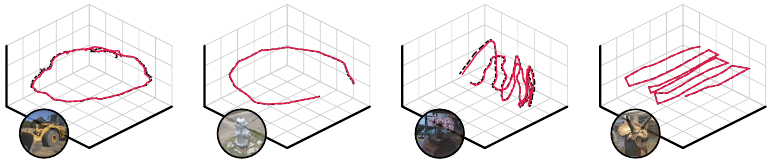}
    \caption{\textbf{Qualitative Pose Estimation Comparison.} FlowMap (solid red) recovers camera poses that are very close to those of COLMAP (dotted black).
    }
    \label{fig:trajectories}
    \vspace{-5pt}
\end{figure*}

%% file: figures/point_clouds.tex
\begin{figure*}[t]
    \centering
    \includegraphics[width=\linewidth]{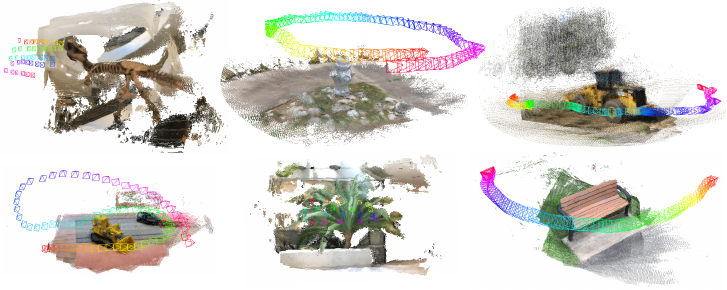}
    \caption{\textbf{Point Clouds Reconstructed by FlowMap.} Unprojecting FlowMap depths using FlowMap's intrinsics and poses yields dense and consistent point clouds.}
    \label{fig:point_clouds}
    \vspace{-10pt}
\end{figure*}

%% file: figures/colmap_comparison.tex
\begin{figure*}[t!]
    \centering
    \includegraphics[width=\linewidth,]{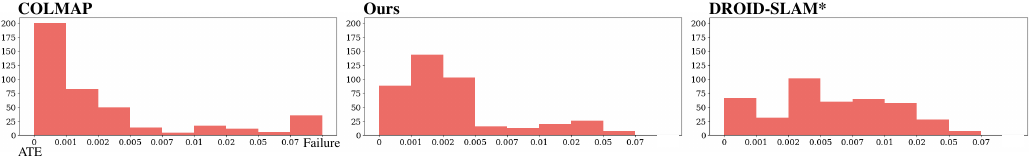}
    \vspace{-20pt}
    \caption{\textbf{Large-scale Robustness Study.}
    We run FlowMap and DROID-SLAM on 420 CO3D scenes across 10 categories and plot mean ATEs with respect to CO3D's COLMAP-generated pose metadata.
    We also re-run COLMAP on the same data.
    Compared to DROID-SLAM, which requires ground-truth intrinsics, FlowMap produces notably lower ATEs.
    FlowMap's ATE distribution is similar to one obtained by re-running COLMAP, with most ATEs falling under 0.005 in both cases.}
    \label{fig:colmap_study}
    \vspace{-5pt}
\end{figure*}

%% file: sec/5_ablations.tex
\input{figures/ablations_table_and_fig}
\section{Ablations and Analysis}
\label{sec:ablations}
We perform ablations to answer the following questions:
\setdefaultleftmargin{.5em}{0em}{}{}{}{}
\begin{itemize}
    \item \underline{Question 1:} Are FlowMap's reparameterizations of depth, pose, and intrinsics necessary, or do free variables perform equally well?
    \item \underline{Question 2:} Are point tracks critical to FlowMap's performance?
    \item \underline{Question 3:} Does self-supervised pre-training of the depth estimation and correspondence weight neural networks improve performance? 
\end{itemize}

\myparagraph{Parameterizations of Depth, Pose, and Camera Intrinsics (Q1)}

We compare the reparameterizations described in Sec.~\ref{sec:reparams} to direct, free-variable optimization of pose, depth, and intrinsics.
Fig.~\ref{fig:ablations} shows qualitative results and quantitative results averaged across 33 scenes.
We find that free-variable variants of FlowMap produce significantly worse reconstruction results and converge much more slowly, confirming that FlowMap's reparameterizations are crucial.

It is worth noting that often, explicitly optimizing a focal length produces high-quality results, as indicated by the relatively high performance of the ``Expl. Focal Length'' ablation.
In fact, given a good initialization, direct focal length regression produces slightly better results than the proposed focal length reparameterization alone on about 80 percent of scenes.
However, on about 20 percent of scenes, this approach falls into a local minimum and reconstruction fails catastrophically.
This justifies the approach FlowMap uses, where the first 1,000 optimization steps use a reparameterized focal length, which is then used to initialize an explicit focal length used for another 1,000 optimization steps.

We further highlight that FlowMap's reparameterizations are necessary to estimate poses and intrinsics in a single forward pass, which is crucial for the generalizable (pre-training) setting explored in Q3.

\myparagraph{Point Tracking (Q2)}
\label{point_track_exp}
While optical flow is only computed between adjacent frames, point track estimators can accurately track points across many frames. 
In Fig.~\ref{fig:ablations}, we show that FlowMap's novel view synthesis performance drops moderately when point tracks are disabled.
Qualitatively, we find that point tracks reduce drift for longer sequences, such as object-centric $360^\circ$ scenes.
This suggests that FlowMap will benefit from further improvements in point tracking methods.
We note that FlowMap's loss formulation is compatible with conventional correspondence methods (e.g. SIFT~\cite{sift} with RANSAC) and learned correspondences~\cite{sarlin20superglue}, which can be treated identically to point tracks.
FlowMap could also be extended to use conventional loop closure mechanisms, which would further reduce drift.

\myparagraph{Pre-training Depth and Correspondence Networks (Q3)}
\input{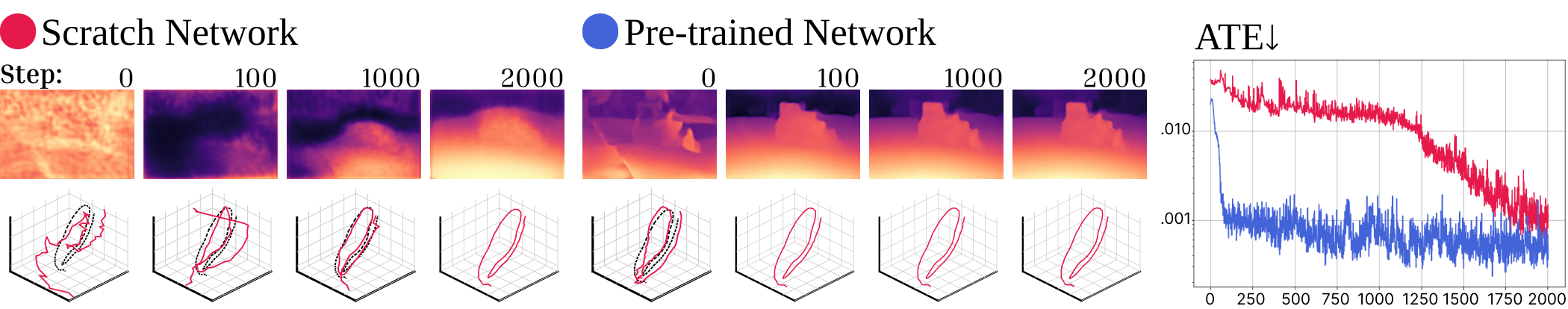}
Since FlowMap is differentiable and provides gradients for any depth-estimating neural network, it is compatible with both randomly initialized neural networks and pre-trained priors.
Learned priors can come from optimization on many scenes, from existing depth estimation models, or from a combination of the two.
In practice, starting with a pre-trained prior leads to significantly faster convergence, as illustrated in Fig.~\ref{fig:prior}.
Note that pre-training and generalization are uniquely enabled by the proposed feed-forward reparameterizations of depth, focal length, and poses.

%% file: figures/ablations_table_and_fig.tex
\begin{figure}[t]
\includegraphics{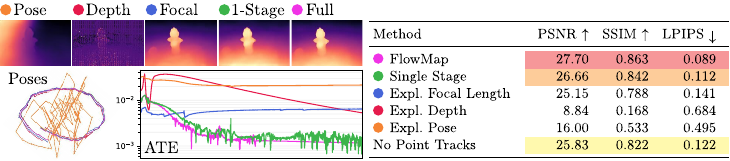}

\caption{
\textbf{Ablations.}
We ablate the proposed feed-forward re-parameterizations of depth, pose, and intrinsics across all datasets.
We find that these reparameterizations are not only critical for high-quality downstream 3D Gaussian Splatting, but also lead to dramatically accelerated convergence, where FlowMap generally converges to high quality poses within a fraction of the optimization steps required for the ablated variants.
We further find that point tracks lead to a significant boost over optical flow alone (right).
See the supplemental document for more ablations.
\vspace{-10pt}
}
\label{fig:ablations}
\end{figure}

%% file: figures/init_vs_pretrain.tex
\begin{figure*}[t!]
    \centering
    \includegraphics[width=\linewidth]{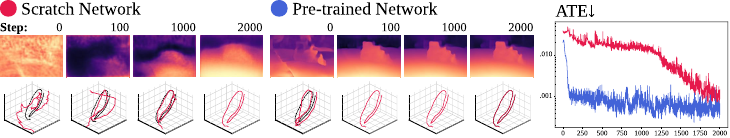}
    \caption{\textbf{Effects of pretraining.} While a randomly initialized FlowMap network often provides accurate poses after optimization, pre-training leads to faster convergence and slightly improved poses. Here we plot depth estimates at specific optimization steps (left) as well as pose accuracy with respect to COLMAP during optimization (right). Randomly initialized FlowMap networks often require more than 20,000 steps to match the accuracy of a pre-trained initialization at 2,000 steps.}
    \label{fig:prior}
\end{figure*}

%% file: sec/6_conclusion.tex
\vspace{-5pt}

\section{Discussion}

\myparagraph{Limitations.}
\label{sec:limitations}
FlowMap has several limitations that suggest exciting directions for future work.
First, FlowMap requires off-the-shelf correspondences from optical flow and point tracking methods. An exciting direction is to remove the dependence on correspondence altogether or to jointly learn correspondence extraction.
Second, we mainly analyze FlowMap in the setting of per-scene optimization, where our results demonstrate that the gradients provided by FlowMap's formulation are robustly lead to high-quality depth and camera parameters. It is natural to attempt to use these gradients to train a feed-forward structure-from-motion method.
Lastly, through its dependence on optical flow or point tracks, FlowMap can currently only process continuous video, in contrast to conventional SfM methods which can operate on unstructured image collections. This is a natural assumption for applications in embodied intelligence, navigation, and robotics, but limits applications in computer graphics. Leveraging unstructured correspondences, e.g. via \cite{sarlin2020superglue}, may be used to overcome this limitation.

\myparagraph{Relationship to Conventional SfM.}
\label{sec:colmap_discussion}
Across many applications of conventional SfM, such as the reconstruction of large, non-continuous image collections, FlowMap cannot serve as a drop-in replacement, and we note that this is not our objective.
Rather, we demonstrate that a self-supervised, end-to-end differentiable, feed-forward formulation that can naturally be integrated into neural network vision models surprisingly approaches COLMAP's performance on the downstream task of novel view synthesis \emph{in the context of video data}.
Here, FlowMap has the potential to make camera pose and depth supervision unnecessary for 3D deep learning, paving the way for training on unannotated, internet-scale video data.

\myparagraph{Conclusion.}
\label{sec:conclusion}
We have introduced FlowMap, a simple, robust, and scalable first-order method for estimating camera parameters from video.
Our model outperforms existing gradient-descent based methods for estimating camera parameters. 
FlowMap's depth and camera parameters enable subsequent reconstruction via Gaussian Splatting of comparable quality to COLMAP.
FlowMap is written in PyTorch and achieves runtimes of 3 minutes for short sequences and 20 minutes for long sequences, and we anticipate that concerted engineering efforts could accelerate FlowMap by an order of magnitude.
Perhaps most excitingly, FlowMap is fully differentiable with respect to per-frame depth estimates.
FlowMap can thus serve as a building block for a new generation of self-supervised monocular depth estimators, deep-learning-based multi-view-geometry methods, and methods for generalizable novel view synthesis~\cite{pixelnerf,tewari2023diffusion,charatan23pixelsplat,wang2021ibrnet,du2023cross,suhail2022generalizable}, unlocking training on internet-scale datasets of unposed videos.

\small\myparagraph{Acknowledgements.} This work was supported by the National Science Foundation under Grant No. 2211259, by the Singapore DSTA under DST00OECI20300823 (New Representations for Vision and 3D Self-Supervised Learning for Label-Efficient Vision), by the Intelligence Advanced Research Projects Activity (IARPA) via Department of Interior/ Interior Business Center (DOI/IBC) under 140D0423C0075, by the Amazon Science Hub, and by IBM. The Toyota Research Institute also partially supported this work. The views and conclusions contained herein reflect the opinions and conclusions of its authors and no other entity. Vincent thanks his Mom for crocheting baby Yoda.

%% file: supplement/additional_ablations.tex
\section{Additional Ablation Studies}
\input{figures/pose_convergence}
\input{tables/ablations_supplemental}
\input{tables/ablations_supplemental_full}
\input{figures/patchmatch_full}
In \cref{tab:ablations_supplemental}, we include three additional ablations.
The ``Random Init.'' ablation uses a randomly initialized CNN for FlowMap training with 2,000 steps of optimization.
The ``Random Init. (20k)'' ablation is identical, but runs for 20,000 optimization steps.
The ``No Corresp. Weights'' ablation removes the correspondence weights used in FlowMap's Procrustes-solving step.
We note that the ``Random Init. (20k)'' ablation's performance almost matches FlowMap's, indicating that although pre-training helps FlowMap converge much more quickly, it is not necessary for accuracy.
In \cref{tab:ablations_supplemental_full}, we report per-scene ablation results.
Finally, \cref{fig:convergence} compares convergence between FlowMap and the free-variable parameterization variants on more scenes.

\input{figures/more_point_clouds}

%% file: figures/pose_convergence.tex
\begin{figure}[!]
    \centering
    \includegraphics[width=\linewidth,]{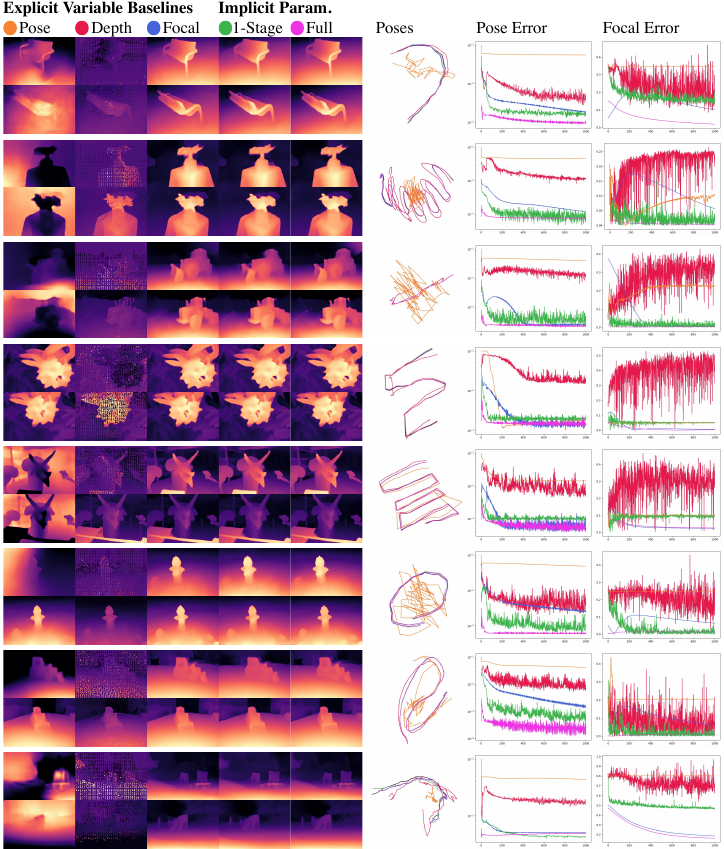}
    \caption{\textbf{Pose and Geometry Convergence for Free-Variable vs. Proposed Parameterizations.}
    We plot poses, depths, focal lengths, and pose error (ATE) obtained with our proposed parameterizations (``Full'') vs. those obtained with free-variable parameterizations at various optimization steps. With our proposed reparameterizations (``Full'') as a baseline, we ablate either depth, focal length, or poses as free-variable optimizations and plot the resulting optimizations' pose and depth estimates. For instance, ``Depth'' corresponds to making the depth an explicit free-variable in the optimization. Using pose-as-variable and depth-as-variable often lead to ``hollow-face'' geometry, where the geometry is effectively inverted but still mostly satisfies the optical flow constraints. We also show results from a single-stage FlowMap pipeline, which only uses the implicit parameterization of intrinsics rather than switching to regressed intrinsics halfway through optimization. Note that the plotted lines for ``Full'' are initialized with the results of ``1-Stage'' and represent the second stage (explicit focal length) of FlowMap optimization.}
    \label{fig:convergence}
\end{figure}

%% file: tables/ablations_supplemental.tex
\setlength{\tabcolsep}{8pt}
\begin{table*}[t]

\centering

\newcommand{\first}{\cellcolor{red!40}}
\newcommand{\second}{\cellcolor{orange!40}}
\newcommand{\third}{\cellcolor{yellow!40}}
\setlength{\tabcolsep}{4pt}

\begin{tabular}{lrrr}
\toprule
Method              & PSNR $\uparrow$ & SSIM $\uparrow$ & LPIPS $\downarrow$ \\
\midrule
FlowMap             &   \first{27.70} &   \first{0.863} &      \first{0.089} \\
Single Stage        &   \third{26.66} &   \third{0.842} &      \third{0.112} \\
Expl. Focal Length  &           25.15 &           0.788 &              0.141 \\
Expl. Depth         &            8.84 &           0.168 &              0.684 \\
Expl. Pose          &           16.00 &           0.533 &              0.495 \\
No Tracks           &           25.83 &           0.822 &              0.122 \\
Random Init.        &           25.54 &           0.808 &              0.129 \\
Random Init. (20k)  &  \second{27.25} &  \second{0.850} &     \second{0.101} \\
No Corresp. Weights &           24.18 &           0.765 &              0.168 \\
\bottomrule
\end{tabular}

\vspace{10pt}

\caption{\textbf{Additional Ablations.} We report additional ablation results averaged across all scenes alongside the ablations found in the main paper.}

\label{tab:ablations_supplemental}
\end{table*}

%% file: tables/ablations_supplemental_full.tex
\setlength{\tabcolsep}{8pt}
\begin{table*}[t]

\newcommand{\first}{\cellcolor{red!40}}
\newcommand{\second}{\cellcolor{orange!40}}
\newcommand{\third}{\cellcolor{yellow!40}}
\setlength{\tabcolsep}{4pt}

\centering
\resizebox{\textwidth}{!}{

\begin{tabular}{l|rrrrr|rrrrr|rrrrr}
\toprule
\multicolumn{1}{c|}{} & \multicolumn{5}{|c|}{Fern (LLFF)} & \multicolumn{5}{|c|}{Flower (LLFF)} & \multicolumn{5}{|c}{Fortress (LLFF)} \\
\midrule
Method              & PSNR $\uparrow$ & SSIM $\uparrow$ & LPIPS $\downarrow$ & Time (min.) $\downarrow$ & ATE     & PSNR $\uparrow$ & SSIM $\uparrow$ & LPIPS $\downarrow$ & Time (min.) $\downarrow$ & ATE     & PSNR $\uparrow$ & SSIM $\uparrow$ & LPIPS $\downarrow$ & Time (min.) $\downarrow$ & ATE     \\
\midrule
FlowMap             &   \first{23.70} &   \first{0.801} &      \first{0.096} &                      4.8 & 0.00233 &           29.07 &   \third{0.877} &      \third{0.084} &                      6.6 & 0.00079 &   \first{31.13} &   \third{0.906} &              0.060 &                      7.8 & 0.00049 \\
Single Stage        &  \second{23.64} &  \second{0.797} &     \second{0.098} &                      4.9 & 0.00294 &           29.06 &   \third{0.877} &              0.086 &                      6.7 & 0.00065 &  \second{31.05} &  \second{0.908} &     \second{0.058} &                      7.9 & 0.00054 \\
Expl. Focal Length  &           23.07 &           0.787 &              0.119 &                      4.6 & 0.00296 &   \third{29.13} &           0.874 &      \first{0.079} &                      6.4 & 0.00293 &           29.82 &           0.891 &              0.062 &                      7.6 & 0.00223 \\
Expl. Depth         &            4.71 &           0.001 &              0.785 &              \first{2.9} & 0.00785 &            8.04 &           0.007 &              0.839 &              \first{3.6} & 0.00666 &            2.60 &           0.001 &              0.774 &              \first{4.1} & 0.00664 \\
Expl. Pose          &            4.71 &           0.001 &              0.785 &                      4.5 & 0.01118 &           15.51 &           0.569 &              0.428 &                      6.3 & 0.00192 &           16.49 &           0.577 &              0.594 &                      7.4 & 0.01302 \\
No Tracks           &   \third{23.58} &   \third{0.796} &      \third{0.099} &             \second{4.3} & 0.00316 &  \second{29.29} &  \second{0.879} &      \third{0.084} &             \second{5.5} & 0.00337 &           30.92 &   \third{0.906} &      \third{0.059} &             \second{6.3} & 0.00143 \\
Random Init.        &           22.68 &           0.756 &              0.113 &                      4.8 & 0.00371 &           28.47 &           0.864 &      \third{0.084} &                      6.6 & 0.00303 &           30.96 &           0.904 &      \third{0.059} &                      7.8 & 0.00068 \\
Random Init. (20k)  &           23.50 &           0.791 &     \second{0.098} &                     44.2 & 0.00312 &   \first{29.33} &   \first{0.880} &     \second{0.083} &                     59.4 & 0.00054 &   \third{31.04} &   \first{0.911} &      \first{0.057} &                     69.7 & 0.00047 \\
No Corresp. Weights &           23.27 &           0.784 &              0.104 &              \third{4.4} & 0.00311 &           27.61 &           0.844 &              0.090 &              \third{6.0} & 0.00554 &           24.05 &           0.709 &              0.138 &              \third{7.1} & 0.01363 \\
\midrule
\multicolumn{1}{c|}{} & \multicolumn{5}{|c|}{Horns (LLFF)} & \multicolumn{5}{|c|}{Orchids (LLFF)} & \multicolumn{5}{|c}{Room (LLFF)} \\
\midrule
Method              & PSNR $\uparrow$ & SSIM $\uparrow$ & LPIPS $\downarrow$ & Time (min.) $\downarrow$ & ATE     & PSNR $\uparrow$ & SSIM $\uparrow$ & LPIPS $\downarrow$ & Time (min.) $\downarrow$ & ATE     & PSNR $\uparrow$ & SSIM $\uparrow$ & LPIPS $\downarrow$ & Time (min.) $\downarrow$ & ATE     \\
\midrule
FlowMap             &   \first{28.35} &   \first{0.903} &     \second{0.071} &                     10.6 & 0.00049 &           19.16 &           0.615 &      \third{0.132} &                      5.5 & 0.00127 &   \first{32.93} &  \second{0.958} &      \first{0.037} &                      7.8 & 0.00274 \\
Single Stage        &           28.19 &   \third{0.899} &     \second{0.071} &                     10.7 & 0.00051 &  \second{19.33} &  \second{0.623} &     \second{0.129} &                      5.6 & 0.00120 &  \second{32.75} &   \first{0.959} &      \first{0.037} &                      7.9 & 0.00265 \\
Expl. Focal Length  &           24.57 &           0.823 &              0.153 &                     10.5 & 0.00100 &           18.96 &           0.606 &              0.151 &                      5.3 & 0.00184 &   \third{32.13} &   \third{0.953} &     \second{0.040} &                      7.6 & 0.00274 \\
Expl. Depth         &            5.94 &           0.002 &              0.788 &              \first{5.3} & 0.00603 &            6.25 &           0.003 &              0.886 &              \first{3.2} & 0.00647 &            5.78 &           0.004 &              0.616 &              \first{4.1} & 0.00710 \\
Expl. Pose          &           14.66 &           0.506 &              0.685 &                     10.0 & 0.01230 &           12.80 &           0.279 &              0.429 &                      5.2 & 0.01496 &           16.92 &           0.767 &              0.466 &                      7.3 & 0.00596 \\
No Tracks           &   \third{28.32} &  \second{0.900} &     \second{0.071} &             \second{8.2} & 0.00173 &   \third{19.21} &   \third{0.616} &              0.133 &             \second{4.8} & 0.00195 &           28.98 &           0.922 &              0.068 &             \second{6.3} & 0.00938 \\
Random Init.        &           23.93 &           0.729 &              0.172 &                     10.6 & 0.00486 &           18.85 &           0.594 &              0.146 &                      5.4 & 0.00188 &           29.19 &           0.920 &              0.067 &                      7.8 & 0.00422 \\
Random Init. (20k)  &  \second{28.33} &  \second{0.900} &      \first{0.068} &                     94.2 & 0.00054 &   \first{19.40} &   \first{0.629} &      \first{0.126} &                     49.5 & 0.00112 &           31.92 &           0.949 &              0.043 &                     69.4 & 0.00288 \\
No Corresp. Weights &           28.17 &           0.893 &      \third{0.072} &              \third{9.6} & 0.00169 &           18.76 &           0.597 &              0.148 &              \third{5.1} & 0.00307 &           31.81 &           0.952 &      \third{0.041} &              \third{7.1} & 0.00331 \\
\midrule
\multicolumn{1}{c|}{} & \multicolumn{5}{|c|}{Trex (LLFF)} & \multicolumn{5}{|c|}{Bonsai (MipNeRF 360)} & \multicolumn{5}{|c}{Kitchen (MipNeRF 360)} \\
\midrule
Method              & PSNR $\uparrow$ & SSIM $\uparrow$ & LPIPS $\downarrow$ & Time (min.) $\downarrow$ & ATE     & PSNR $\uparrow$ & SSIM $\uparrow$ & LPIPS $\downarrow$ & Time (min.) $\downarrow$ & ATE     & PSNR $\uparrow$ & SSIM $\uparrow$ & LPIPS $\downarrow$ & Time (min.) $\downarrow$ & ATE     \\
\midrule
FlowMap             &   \third{26.27} &   \third{0.880} &     \second{0.075} &                      9.7 & 0.00655 &   \first{32.24} &  \second{0.950} &      \first{0.047} &                     24.2 & 0.00048 &  \second{30.47} &  \second{0.936} &     \second{0.049} &                     10.9 & 0.00041 \\
Single Stage        &   \first{26.65} &   \first{0.886} &      \first{0.073} &                      9.8 & 0.00655 &  \second{32.21} &   \first{0.951} &     \second{0.048} &                     24.3 & 0.00046 &   \third{30.26} &   \third{0.925} &      \third{0.051} &                     11.1 & 0.00072 \\
Expl. Focal Length  &           24.57 &           0.852 &              0.107 &                      9.5 & 0.00766 &           21.36 &           0.689 &              0.231 &                     24.1 & 0.00184 &           21.29 &           0.645 &              0.174 &                     10.7 & 0.00449 \\
Expl. Depth         &            5.56 &           0.002 &              0.759 &              \first{4.8} & 0.04406 &           10.46 &           0.045 &              0.633 &             \first{11.3} & 0.02830 &            5.08 &           0.016 &              0.753 &              \first{5.3} & 0.00827 \\
Expl. Pose          &           15.09 &           0.540 &              0.544 &                      9.2 & 0.02277 &           13.12 &           0.425 &              0.577 &                     22.8 & 0.01407 &           14.18 &           0.387 &              0.587 &                     10.3 & 0.01669 \\
No Tracks           &           24.36 &           0.831 &              0.110 &             \second{7.8} & 0.02011 &           25.53 &   \third{0.863} &      \third{0.115} &            \second{18.0} & 0.00291 &           25.48 &           0.794 &              0.112 &             \second{8.5} & 0.00302 \\
Random Init.        &           24.84 &           0.835 &              0.099 &                      9.7 & 0.00598 &           18.38 &           0.585 &              0.342 &                     24.2 & 0.01380 &           24.94 &           0.764 &              0.113 &                     10.9 & 0.00345 \\
Random Init. (20k)  &  \second{26.45} &  \second{0.882} &      \third{0.076} &                     86.5 & 0.00991 &           18.75 &           0.600 &              0.342 &                    214.8 & 0.01433 &   \first{31.69} &   \first{0.945} &      \first{0.044} &                     96.8 & 0.00023 \\
No Corresp. Weights &           25.33 &           0.859 &              0.094 &              \third{8.8} & 0.01083 &   \third{25.59} &           0.841 &              0.118 &             \third{21.8} & 0.00141 &           24.62 &           0.742 &              0.118 &              \third{9.9} & 0.00422 \\
\midrule
\multicolumn{1}{c|}{} & \multicolumn{5}{|c|}{Counter (MipNeRF 360)} & \multicolumn{5}{|c|}{Barn (Tanks \& Temples)} & \multicolumn{5}{|c}{Caterpillar (Tanks \& Temples)} \\
\midrule
Method              & PSNR $\uparrow$ & SSIM $\uparrow$ & LPIPS $\downarrow$ & Time (min.) $\downarrow$ & ATE     & PSNR $\uparrow$ & SSIM $\uparrow$ & LPIPS $\downarrow$ & Time (min.) $\downarrow$ & ATE     & PSNR $\uparrow$ & SSIM $\uparrow$ & LPIPS $\downarrow$ & Time (min.) $\downarrow$ & ATE     \\
\midrule
FlowMap             &  \second{26.80} &  \second{0.862} &     \second{0.121} &                     24.2 & 0.00076 &   \first{27.10} &  \second{0.872} &     \second{0.090} &                     22.3 & 0.00048 &  \second{28.25} &  \second{0.830} &     \second{0.113} &                     22.3 & 0.00030 \\
Single Stage        &   \third{26.65} &   \third{0.857} &      \third{0.124} &                     24.3 & 0.00083 &           26.28 &           0.857 &              0.104 &                     22.4 & 0.00102 &   \first{28.32} &   \first{0.834} &      \first{0.110} &                     22.4 & 0.00026 \\
Expl. Focal Length  &           21.82 &           0.697 &              0.237 &                     24.1 & 0.00355 &  \second{27.04} &   \first{0.873} &      \first{0.086} &                     21.9 & 0.00085 &           27.12 &           0.789 &              0.133 &                     22.1 & 0.00046 \\
Expl. Depth         &            8.80 &           0.029 &              0.719 &             \first{11.4} & 0.01003 &           17.01 &           0.591 &              0.489 &             \first{10.7} & 0.00923 &            9.15 &           0.016 &              0.732 &             \first{10.7} & 0.00841 \\
Expl. Pose          &           14.40 &           0.506 &              0.554 &                     22.9 & 0.00611 &           18.09 &           0.625 &              0.455 &                     20.9 & 0.02160 &           17.57 &           0.491 &              0.554 &                     21.1 & 0.00817 \\
No Tracks           &           23.91 &           0.788 &              0.183 &            \second{18.1} & 0.00240 &           25.41 &           0.837 &              0.122 &            \second{16.1} & 0.00363 &           27.33 &           0.807 &              0.133 &            \second{16.1} & 0.00095 \\
Random Init.        &           26.05 &           0.847 &              0.131 &                     24.2 & 0.00088 &           26.24 &           0.864 &              0.100 &                     22.2 & 0.00079 &           26.27 &           0.750 &              0.169 &                     22.2 & 0.00147 \\
Random Init. (20k)  &   \first{26.88} &   \first{0.867} &      \first{0.115} &                    214.3 & 0.00064 &   \third{26.80} &   \third{0.871} &      \third{0.091} &                    197.3 & 0.00049 &   \third{28.01} &   \third{0.823} &      \third{0.122} &                    197.3 & 0.00031 \\
No Corresp. Weights &           17.93 &           0.575 &              0.391 &             \third{21.8} & 0.01099 &           24.53 &           0.820 &              0.133 &             \third{20.3} & 0.00244 &           25.93 &           0.734 &              0.174 &             \third{20.1} & 0.00106 \\
\midrule
\multicolumn{1}{c|}{} & \multicolumn{5}{|c|}{Church (Tanks \& Temples)} & \multicolumn{5}{|c|}{Courthouse (Tanks \& Temples)} & \multicolumn{5}{|c}{Family (Tanks \& Temples)} \\
\midrule
Method              & PSNR $\uparrow$ & SSIM $\uparrow$ & LPIPS $\downarrow$ & Time (min.) $\downarrow$ & ATE     & PSNR $\uparrow$ & SSIM $\uparrow$ & LPIPS $\downarrow$ & Time (min.) $\downarrow$ & ATE     & PSNR $\uparrow$ & SSIM $\uparrow$ & LPIPS $\downarrow$ & Time (min.) $\downarrow$ & ATE     \\
\midrule
FlowMap             &  \second{28.29} &  \second{0.883} &      \first{0.074} &                     22.4 & 0.00061 &   \third{27.51} &  \second{0.911} &     \second{0.055} &                     22.2 & 0.00129 &  \second{27.96} &  \second{0.889} &     \second{0.067} &                     22.1 & 0.00039 \\
Single Stage        &           27.60 &   \third{0.875} &     \second{0.079} &                     22.3 & 0.00061 &   \first{27.67} &   \first{0.914} &      \first{0.054} &                     22.3 & 0.00150 &   \third{27.65} &   \third{0.880} &      \third{0.075} &                     22.4 & 0.00033 \\
Expl. Focal Length  &           27.29 &           0.856 &              0.088 &                     22.0 & 0.00093 &           26.86 &   \third{0.897} &      \third{0.069} &                     22.1 & 0.00234 &           27.10 &           0.873 &              0.082 &                     22.0 & 0.00082 \\
Expl. Depth         &           16.21 &           0.518 &              0.474 &             \first{10.7} & 0.02314 &            3.52 &           0.001 &              0.745 &             \first{10.7} & 0.00718 &            4.09 &           0.001 &              0.773 &             \first{10.8} & 0.01403 \\
Expl. Pose          &           17.66 &           0.582 &              0.457 &                     21.0 & 0.00807 &           19.68 &           0.726 &              0.251 &                     21.0 & 0.00511 &           15.79 &           0.562 &              0.507 &                     21.1 & 0.03074 \\
No Tracks           &           26.93 &           0.851 &              0.100 &            \second{16.1} & 0.00259 &           25.27 &           0.858 &              0.108 &            \second{16.1} & 0.00442 &           27.00 &           0.869 &              0.088 &            \second{16.1} & 0.00172 \\
Random Init.        &           27.45 &           0.858 &              0.089 &                     22.2 & 0.00112 &           26.55 &           0.894 &              0.071 &                     22.2 & 0.00314 &           26.36 &           0.858 &              0.093 &                     22.2 & 0.00148 \\
Random Init. (20k)  &   \first{28.67} &   \first{0.886} &      \first{0.074} &                    197.4 & 0.00030 &  \second{27.62} &  \second{0.911} &      \first{0.054} &                    197.8 & 0.00101 &   \first{28.07} &   \first{0.892} &      \first{0.066} &                    196.9 & 0.00019 \\
No Corresp. Weights &   \third{27.86} &   \third{0.875} &      \third{0.081} &             \third{20.3} & 0.00086 &           25.62 &           0.868 &              0.086 &             \third{20.1} & 0.00264 &           19.01 &           0.629 &              0.313 &             \third{20.2} & 0.00672 \\
\midrule
\multicolumn{1}{c|}{} & \multicolumn{5}{|c|}{Francis (Tanks \& Temples)} & \multicolumn{5}{|c|}{Horse (Tanks \& Temples)} & \multicolumn{5}{|c}{Ignatius (Tanks \& Temples)} \\
\midrule
Method              & PSNR $\uparrow$ & SSIM $\uparrow$ & LPIPS $\downarrow$ & Time (min.) $\downarrow$ & ATE     & PSNR $\uparrow$ & SSIM $\uparrow$ & LPIPS $\downarrow$ & Time (min.) $\downarrow$ & ATE     & PSNR $\uparrow$ & SSIM $\uparrow$ & LPIPS $\downarrow$ & Time (min.) $\downarrow$ & ATE     \\
\midrule
FlowMap             &  \second{31.90} &  \second{0.903} &     \second{0.080} &                     22.4 & 0.00058 &   \first{28.35} &   \first{0.917} &      \first{0.064} &                     22.4 & 0.00054 &   \first{24.54} &   \first{0.773} &      \first{0.131} &                     22.4 & 0.00037 \\
Single Stage        &   \first{32.20} &   \first{0.905} &      \first{0.078} &                     22.4 & 0.00054 &           11.42 &           0.635 &              0.496 &                     22.3 & 0.03959 &  \second{24.48} &   \third{0.764} &      \first{0.131} &                     22.4 & 0.00024 \\
Expl. Focal Length  &           30.89 &           0.884 &              0.108 &                     22.0 & 0.00040 &   \third{27.82} &   \third{0.905} &      \third{0.074} &                     22.0 & 0.00102 &           23.12 &           0.723 &      \third{0.157} &                     22.0 & 0.00071 \\
Expl. Depth         &            7.42 &           0.006 &              0.631 &             \first{10.8} & 0.01956 &            2.67 &           0.000 &              0.691 &             \first{10.7} & 0.02555 &            5.68 &           0.006 &              0.867 &             \first{10.7} & 0.02181 \\
Expl. Pose          &           18.19 &           0.639 &              0.464 &                     20.9 & 0.03102 &           14.60 &           0.661 &              0.468 &                     21.0 & 0.03918 &           12.48 &           0.314 &              0.640 &                     20.9 & 0.02886 \\
No Tracks           &           30.72 &           0.887 &              0.100 &            \second{16.1} & 0.00113 &           25.50 &           0.882 &              0.101 &            \second{16.1} & 0.00241 &           23.54 &           0.727 &              0.163 &            \second{16.1} & 0.00144 \\
Random Init.        &           29.44 &           0.862 &              0.122 &                     22.3 & 0.00289 &           25.07 &           0.871 &              0.119 &                     22.2 & 0.00380 &           23.50 &           0.737 &              0.159 &                     22.1 & 0.00084 \\
Random Init. (20k)  &   \third{31.56} &   \third{0.899} &      \third{0.085} &                    197.7 & 0.00138 &  \second{28.16} &  \second{0.915} &     \second{0.067} &                    197.0 & 0.00066 &   \third{24.47} &  \second{0.771} &     \second{0.133} &                    197.5 & 0.00034 \\
No Corresp. Weights &           28.92 &           0.850 &              0.130 &             \third{20.1} & 0.00397 &           25.82 &           0.871 &              0.100 &             \third{20.2} & 0.00275 &           21.89 &           0.655 &              0.197 &             \third{20.2} & 0.00108 \\
\midrule
\multicolumn{1}{c|}{} & \multicolumn{5}{|c|}{M60 (Tanks \& Temples)} & \multicolumn{5}{|c|}{Museum (Tanks \& Temples)} & \multicolumn{5}{|c}{Panther (Tanks \& Temples)} \\
\midrule
Method              & PSNR $\uparrow$ & SSIM $\uparrow$ & LPIPS $\downarrow$ & Time (min.) $\downarrow$ & ATE     & PSNR $\uparrow$ & SSIM $\uparrow$ & LPIPS $\downarrow$ & Time (min.) $\downarrow$ & ATE     & PSNR $\uparrow$ & SSIM $\uparrow$ & LPIPS $\downarrow$ & Time (min.) $\downarrow$ & ATE     \\
\midrule
FlowMap             &   \third{23.23} &  \second{0.805} &      \third{0.190} &                     22.4 & 0.00838 &   \third{28.48} &   \third{0.862} &      \third{0.078} &                     22.2 & 0.00070 &  \second{27.50} &  \second{0.882} &     \second{0.105} &                     22.3 & 0.00112 \\
Single Stage        &  \second{23.30} &   \third{0.803} &     \second{0.187} &                     22.3 & 0.00832 &  \second{28.62} &  \second{0.866} &     \second{0.076} &                     22.4 & 0.00058 &   \third{27.31} &   \third{0.881} &      \first{0.104} &                     22.4 & 0.00118 \\
Expl. Focal Length  &           19.65 &           0.696 &              0.278 &                     22.1 & 0.01400 &           28.15 &           0.850 &              0.092 &                     22.0 & 0.00124 &           21.86 &           0.737 &              0.239 &                     22.1 & 0.00893 \\
Expl. Depth         &           13.65 &           0.529 &              0.547 &             \first{10.7} & 0.01945 &           16.55 &           0.507 &              0.489 &             \first{10.7} & 0.03144 &           16.17 &           0.619 &              0.456 &             \first{10.5} & 0.01527 \\
Expl. Pose          &           14.37 &           0.566 &              0.546 &                     21.1 & 0.01454 &           16.12 &           0.489 &              0.524 &                     21.1 & 0.03128 &           16.41 &           0.613 &              0.460 &                     20.9 & 0.00523 \\
No Tracks           &           23.17 &  \second{0.805} &              0.195 &            \second{16.1} & 0.00674 &           27.63 &           0.844 &              0.096 &            \second{16.1} & 0.00150 &           24.69 &           0.833 &              0.161 &            \second{16.1} & 0.00605 \\
Random Init.        &           21.81 &           0.781 &              0.206 &                     22.3 & 0.01008 &           27.57 &           0.849 &              0.094 &                     22.3 & 0.00108 &           25.25 &           0.845 &              0.141 &                     22.2 & 0.00352 \\
Random Init. (20k)  &   \first{23.61} &   \first{0.817} &      \first{0.171} &                    197.7 & 0.00869 &   \first{28.74} &   \first{0.868} &      \first{0.075} &                    196.9 & 0.00067 &   \first{27.61} &   \first{0.884} &     \second{0.105} &                    198.1 & 0.00131 \\
No Corresp. Weights &            9.44 &           0.545 &              0.651 &             \third{20.2} & 0.02315 &           27.94 &           0.849 &              0.086 &             \third{20.1} & 0.00122 &           26.00 &           0.850 &      \third{0.138} &             \third{20.0} & 0.00182 \\
\midrule
\multicolumn{1}{c|}{} & \multicolumn{5}{|c|}{Playground (Tanks \& Temples)} & \multicolumn{5}{|c|}{Train (Tanks \& Temples)} & \multicolumn{5}{|c}{Truck (Tanks \& Temples)} \\
\midrule
Method              & PSNR $\uparrow$ & SSIM $\uparrow$ & LPIPS $\downarrow$ & Time (min.) $\downarrow$ & ATE     & PSNR $\uparrow$ & SSIM $\uparrow$ & LPIPS $\downarrow$ & Time (min.) $\downarrow$ & ATE     & PSNR $\uparrow$ & SSIM $\uparrow$ & LPIPS $\downarrow$ & Time (min.) $\downarrow$ & ATE     \\
\midrule
FlowMap             &   \first{24.29} &   \first{0.727} &      \first{0.192} &                     22.2 & 0.00096 &  \second{26.22} &  \second{0.870} &     \second{0.077} &                     22.2 & 0.00082 &   \first{24.34} &   \first{0.828} &      \first{0.098} &                     22.3 & 0.00078 \\
Single Stage        &  \second{23.39} &  \second{0.710} &     \second{0.209} &                     22.4 & 0.00102 &   \third{26.11} &   \third{0.863} &      \third{0.082} &                     22.3 & 0.00100 &           24.21 &  \second{0.826} &     \second{0.102} &                     22.3 & 0.00074 \\
Expl. Focal Length  &           18.55 &           0.525 &              0.380 &                     21.9 & 0.00621 &           25.82 &           0.847 &              0.092 &                     22.1 & 0.00199 &           24.03 &           0.816 &              0.110 &                     22.0 & 0.00070 \\
Expl. Depth         &           14.29 &           0.427 &              0.637 &             \first{10.6} & 0.01642 &           15.89 &           0.553 &              0.464 &             \first{10.7} & 0.01233 &           13.50 &           0.477 &              0.621 &             \first{10.7} & 0.01430 \\
Expl. Pose          &           15.78 &           0.467 &              0.625 &                     21.0 & 0.03699 &           20.86 &           0.713 &              0.228 &                     21.0 & 0.01469 &           14.39 &           0.509 &              0.556 &                     21.0 & 0.01850 \\
No Tracks           &   \third{22.29} &           0.681 &              0.267 &            \second{16.1} & 0.00215 &           25.75 &           0.856 &              0.087 &            \second{16.2} & 0.00131 &           23.32 &           0.786 &              0.150 &            \second{16.1} & 0.00221 \\
Random Init.        &           21.78 &           0.654 &              0.262 &                     22.1 & 0.00315 &           22.80 &           0.769 &              0.163 &                     22.2 & 0.01824 &   \third{24.26} &   \third{0.823} &      \third{0.104} &                     22.2 & 0.00064 \\
Random Init. (20k)  &  \second{23.39} &   \third{0.701} &      \third{0.229} &                    196.8 & 0.00092 &   \first{26.66} &   \first{0.877} &      \first{0.072} &                    197.4 & 0.00089 &  \second{24.28} &           0.816 &              0.109 &                    197.7 & 0.00100 \\
No Corresp. Weights &           17.10 &           0.477 &              0.456 &             \third{20.2} & 0.03438 &           23.22 &           0.780 &              0.138 &             \third{20.0} & 0.00283 &           23.18 &           0.792 &              0.122 &             \third{20.1} & 0.00109 \\
\midrule
\multicolumn{1}{c|}{} & \multicolumn{5}{|c|}{Bench (CO3D)} & \multicolumn{5}{|c}{Hydrant (CO3D)} \\
\midrule
Method              & PSNR $\uparrow$ & SSIM $\uparrow$ & LPIPS $\downarrow$ & Time (min.) $\downarrow$ & ATE     & PSNR $\uparrow$ & SSIM $\uparrow$ & LPIPS $\downarrow$ & Time (min.) $\downarrow$ & ATE     \\
\midrule
FlowMap             &  \second{33.17} &  \second{0.927} &     \second{0.045} &                     22.0 & 0.03094 &           29.05 &           0.865 &              0.083 &                     22.1 & 0.00083 \\
Single Stage        &           25.13 &           0.701 &              0.174 &                     22.2 & 0.03539 &  \second{29.58} &  \second{0.884} &     \second{0.073} &                     22.2 & 0.00094 \\
Expl. Focal Length  &           29.06 &           0.851 &              0.096 &                     21.8 & 0.03244 &           22.70 &           0.543 &              0.230 &                     21.9 & 0.00636 \\
Expl. Depth         &            5.41 &           0.001 &              0.819 &             \first{10.5} & 0.02829 &            5.31 &           0.000 &              0.789 &             \first{10.5} & 0.00533 \\
Expl. Pose          &           21.84 &           0.638 &              0.312 &                     20.8 & 0.03456 &           24.39 &           0.699 &              0.188 &                     20.8 & 0.00650 \\
No Tracks           &           29.18 &           0.861 &              0.085 &            \second{16.0} & 0.03298 &           24.39 &           0.692 &              0.173 &            \second{16.0} & 0.00679 \\
Random Init.        &   \third{32.27} &   \third{0.914} &      \third{0.054} &                     22.1 & 0.03100 &   \third{29.16} &   \third{0.874} &      \third{0.080} &                     22.1 & 0.00076 \\
Random Init. (20k)  &   \first{33.45} &   \first{0.931} &      \first{0.044} &                    196.6 & 0.03068 &   \first{30.09} &   \first{0.896} &      \first{0.065} &                    197.7 & 0.00044 \\
No Corresp. Weights &           31.54 &           0.902 &              0.062 &             \third{20.0} & 0.03212 &           23.55 &           0.596 &              0.193 &             \third{20.0} & 0.00553 \\
\bottomrule
\end{tabular}

}

\vspace{5pt}
\caption{Ablations for all individual scenes on all datasets.}
\label{tab:ablations_supplemental_full}
\end{table*}

%% file: figures/patchmatch_full.tex
\begin{figure*}[H]
    \centering
    \includegraphics[width=\linewidth,]{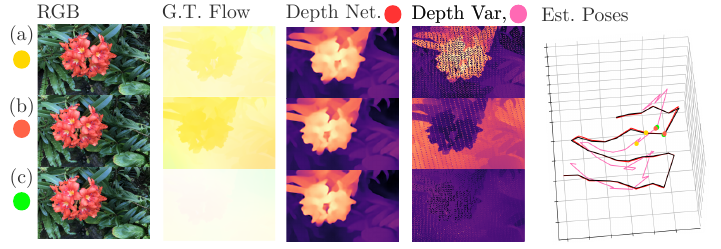}
     \caption{\textbf{PatchMatch.} Frame (a) and (b) both contain geometry-informative flow, but due to the ambiguity of optical flow relating depth and speed of motion, different frames in the depth-variable optimization converge to different solutions; using a depth CNN yields a consistent solution to this ambiguity across frames. And in the case of small or rotation-dominant motion (c), the flow does not sufficiently inform geometry and the optimized depth is mostly planar.}
  \label{fig:patchmatch}
\end{figure*}

%% file: figures/more_point_clouds.tex
\begin{figure}[th]
    \centering
    \includegraphics[width=\linewidth,]{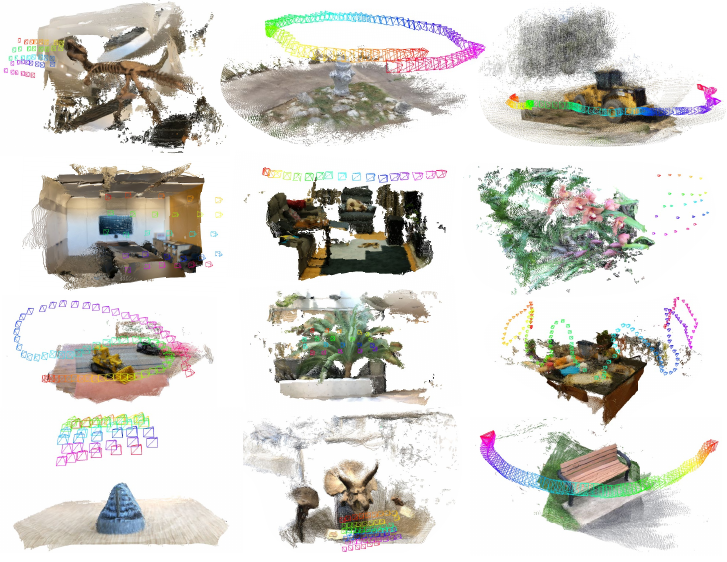}
    \caption{\textbf{Additional Point Clouds} Here we plot additional point clouds across the Tanks and Temples, LLFF, Mip-NeRF360, and CO3D datasets.}
    \label{fig:add_clouds}
\end{figure}

%% file: supplement/additional_results.tex
\section{Additional Results}

\input{tables/table_main_comparison_supplemental}

\subsection{Pre-Trained Depth vs. Fine-Tuned Depth vs. High-Resolution Fine-Tuned Depth.}
In \cref{fig:depths}, we compare the depths produced by FlowMap's initialization to the depths produced after FlowMap optimization.
We additionally compare these results to a MiDaS CNN fine-tuned at a significantly higher resolution.
We find that per-scene fine-tuning leads to high-quality depth predictions.
This is illustrated by Fig.~\ref{fig:add_clouds}, which demonstrates FlowMap's ability to generate high-quality, consistent depths.
However, it is worth noting that FlowMap's off-the-shelf depths are slightly blurry.
To investigate whether this is a limitation of our loss or the architecture of the depth-predicting CNN, we also perform optimization at a higher resolution.
We find that this leads to crisp depth maps, demonstrating that blurry depth maps are a result of insufficient capacity of the MiDaS backbone and not a limitation of our camera-induced flow loss.
Notably, the poses barely change in this fine-tuning stage.
It is likely that replacing the MiDaS depth predictor with a more powerful depth backbone would lead to sharper depth without high-resolution fine-tuning.

\input{figures/depth_vis}

\input{figures/patchmatch_full}
\subsection{Additional Point Clouds and Qualitative Pose Reconstructions}
In \cref{fig:add_clouds}, we display 12 additional point clouds plus estimated camera poses across popular datasets and scenes across the LLFF, Tanks and Temples, MipNeRF 360, and CO3D datasets.
FlowMap robustly recovers camera poses and scene geometry across these diverse, challenging, and real-world sequences.

\subsection{Failure Cases}
While running FlowMap, we observed failures on several scenes.
These include the Tanks-and-Temples Auditorium scene (our model struggles with rotation-dominant trajectories), the LLFF Leaves scene (our model falls into a ``hollow-face minimum''), and the Tanks-and-Temples Lighthouse scene (this video features a large lens flare which degrades the optical flow).
Future extensions to FlowMap could use an occlusion-aware formulation to avoid hollow-face minima.

%% file: tables/table_main_comparison_supplemental.tex
\setlength{\tabcolsep}{8pt}
\begin{table*}[t]

\newcommand{\first}{\cellcolor{red!40}}
\newcommand{\second}{\cellcolor{orange!40}}
\newcommand{\third}{\cellcolor{yellow!40}}
\setlength{\tabcolsep}{4pt}

\centering
\resizebox{\textwidth}{!}{

\begin{tabular}{l|rrrrr|rrrrr|rrrrr}
\toprule
\multicolumn{1}{c|}{} & \multicolumn{5}{|c|}{Fern (LLFF)} & \multicolumn{5}{|c|}{Flower (LLFF)} & \multicolumn{5}{|c}{Fortress (LLFF)} \\
\midrule
Method       & PSNR $\uparrow$ & SSIM $\uparrow$ & LPIPS $\downarrow$ & Time (min.) $\downarrow$ & ATE     & PSNR $\uparrow$ & SSIM $\uparrow$ & LPIPS $\downarrow$ & Time (min.) $\downarrow$ & ATE     & PSNR $\uparrow$ & SSIM $\uparrow$ & LPIPS $\downarrow$ & Time (min.) $\downarrow$ & ATE     \\
\midrule
FlowMap      &   \third{23.70} &   \third{0.801} &     \second{0.096} &              \third{4.8} & 0.00233 &   \third{29.07} &   \third{0.877} &     \second{0.084} &              \third{6.6} & 0.00079 &   \first{31.13} &  \second{0.906} &     \second{0.060} &              \third{7.8} & 0.00049 \\
COLMAP       &  \second{24.04} &  \second{0.818} &              0.133 &             \second{0.4} &     N/A &  \second{29.60} &  \second{0.884} &              0.090 &             \second{2.5} &     N/A &           25.69 &   \third{0.892} &              0.087 &             \second{1.4} &     N/A \\
COLMAP (MVS) &   \first{24.33} &   \first{0.826} &      \first{0.094} &                      6.7 &     N/A &   \first{29.82} &   \first{0.888} &      \third{0.085} &                     11.3 &     N/A &  \second{30.97} &   \first{0.909} &      \first{0.059} &                     13.9 &     N/A \\
DROID-SLAM*  &           23.13 &           0.752 &      \third{0.125} &              \first{0.1} & 0.00089 &           28.48 &           0.860 &      \first{0.079} &              \first{0.2} & 0.00162 &   \third{30.05} &           0.856 &      \third{0.065} &              \first{0.3} & 0.00038 \\
NoPE-NeRF*   &           19.33 &           0.520 &              0.580 &                   1227.0 & 0.01470 &           19.63 &           0.540 &              0.470 &                   1777.7 & 0.02581 &           21.00 &           0.530 &              0.510 &                    533.4 & 0.02068 \\
\midrule
\multicolumn{1}{c|}{} & \multicolumn{5}{|c|}{Horns (LLFF)} & \multicolumn{5}{|c|}{Orchids (LLFF)} & \multicolumn{5}{|c}{Room (LLFF)} \\
\midrule
Method       & PSNR $\uparrow$ & SSIM $\uparrow$ & LPIPS $\downarrow$ & Time (min.) $\downarrow$ & ATE     & PSNR $\uparrow$ & SSIM $\uparrow$ & LPIPS $\downarrow$ & Time (min.) $\downarrow$ & ATE     & PSNR $\uparrow$ & SSIM $\uparrow$ & LPIPS $\downarrow$ & Time (min.) $\downarrow$ & ATE     \\
\midrule
FlowMap      &   \third{28.35} &   \first{0.903} &      \third{0.071} &             \third{10.6} & 0.00049 &   \third{19.16} &   \third{0.615} &      \third{0.132} &              \third{5.5} & 0.00127 &  \second{32.93} &  \second{0.958} &     \second{0.037} &             \second{7.8} & 0.00274 \\
COLMAP       &           27.82 &   \third{0.888} &              0.095 &             \second{1.5} &     N/A &  \second{19.33} &  \second{0.636} &     \second{0.126} &             \second{0.7} &     N/A &           25.69 &   \third{0.927} &              0.096 &              \first{0.3} &     N/A \\
COLMAP (MVS) &   \first{28.68} &  \second{0.902} &     \second{0.067} &                     20.5 &     N/A &   \first{19.79} &   \first{0.657} &      \first{0.117} &                      8.7 &     N/A &   \first{33.43} &   \first{0.963} &      \first{0.035} &             \third{14.6} &     N/A \\
DROID-SLAM*  &  \second{28.37} &           0.881 &      \first{0.064} &              \first{0.5} & 0.00045 &           18.44 &           0.555 &              0.179 &              \first{0.2} & 0.00072 &   \third{27.63} &           0.924 &      \third{0.078} &              \first{0.3} & 0.00051 \\
NoPE-NeRF*   &           11.88 &           0.370 &              0.820 &                   2597.7 & 0.07315 &           13.11 &           0.270 &              0.620 &                   1377.9 & 0.05492 &           17.79 &           0.650 &              0.590 &                   2500.5 & 0.03714 \\
\midrule
\multicolumn{1}{c|}{} & \multicolumn{5}{|c|}{Trex (LLFF)} & \multicolumn{5}{|c|}{Bonsai (MipNeRF 360)} & \multicolumn{5}{|c}{Kitchen (MipNeRF 360)} \\
\midrule
Method       & PSNR $\uparrow$ & SSIM $\uparrow$ & LPIPS $\downarrow$ & Time (min.) $\downarrow$ & ATE     & PSNR $\uparrow$ & SSIM $\uparrow$ & LPIPS $\downarrow$ & Time (min.) $\downarrow$ & ATE     & PSNR $\uparrow$ & SSIM $\uparrow$ & LPIPS $\downarrow$ & Time (min.) $\downarrow$ & ATE     \\
\midrule
FlowMap      &           26.27 &           0.880 &              0.075 &              \third{9.7} & 0.00655 &   \third{32.24} &  \second{0.950} &     \second{0.047} &             \third{24.2} & 0.00048 &  \second{30.47} &  \second{0.936} &     \second{0.049} &             \third{10.9} & 0.00041 \\
COLMAP       &  \second{27.95} &  \second{0.912} &     \second{0.062} &             \second{1.1} &     N/A &  \second{32.64} &   \third{0.949} &      \third{0.058} &             \second{6.9} &     N/A &           28.82 &  \second{0.936} &              0.056 &             \second{3.4} &     N/A \\
COLMAP (MVS) &   \first{28.92} &   \first{0.922} &      \first{0.049} &                     18.4 &     N/A &   \first{33.14} &   \first{0.957} &      \first{0.045} &                     52.2 &     N/A &   \first{31.33} &   \first{0.948} &      \first{0.045} &                     22.4 &     N/A \\
DROID-SLAM*  &   \third{27.36} &   \third{0.898} &      \third{0.067} &              \first{0.3} & 0.00062 &           31.96 &           0.947 &      \first{0.045} &              \first{0.9} & 0.00016 &   \third{29.75} &   \third{0.903} &      \third{0.054} &              \first{0.4} & 0.00015 \\
NoPE-NeRF*   &           18.71 &           0.550 &              0.550 &                   2614.1 & 0.04796 &           13.49 &           0.370 &              0.770 &                   2615.2 & 0.04475 &           14.86 &           0.370 &              0.710 &                    516.3 & 0.05471 \\
\midrule
\multicolumn{1}{c|}{} & \multicolumn{5}{|c|}{Counter (MipNeRF 360)} & \multicolumn{5}{|c|}{Barn (Tanks \& Temples)} & \multicolumn{5}{|c}{Caterpillar (Tanks \& Temples)} \\
\midrule
Method       & PSNR $\uparrow$ & SSIM $\uparrow$ & LPIPS $\downarrow$ & Time (min.) $\downarrow$ & ATE     & PSNR $\uparrow$ & SSIM $\uparrow$ & LPIPS $\downarrow$ & Time (min.) $\downarrow$ & ATE     & PSNR $\uparrow$ & SSIM $\uparrow$ & LPIPS $\downarrow$ & Time (min.) $\downarrow$ & ATE     \\
\midrule
FlowMap      &           26.80 &           0.862 &              0.121 &             \third{24.2} & 0.00076 &   \third{27.10} &           0.872 &      \third{0.090} &             \third{22.3} & 0.00048 &  \second{28.25} &  \second{0.830} &      \third{0.113} &             \third{22.3} & 0.00030 \\
COLMAP       &  \second{28.39} &  \second{0.899} &      \third{0.107} &             \second{4.1} &     N/A &  \second{27.18} &   \third{0.874} &              0.108 &             \second{3.5} &     N/A &           28.05 &           0.825 &              0.134 &             \second{6.6} &     N/A \\
COLMAP (MVS) &   \first{28.61} &   \first{0.909} &      \first{0.089} &                     52.9 &     N/A &   \first{27.91} &   \first{0.889} &      \first{0.075} &                     51.5 &     N/A &   \first{28.52} &   \first{0.839} &      \first{0.103} &                     51.1 &     N/A \\
DROID-SLAM*  &   \third{27.78} &   \third{0.890} &     \second{0.099} &              \first{0.7} & 0.00019 &           27.03 &  \second{0.877} &     \second{0.082} &              \first{0.8} & 0.00029 &   \third{28.13} &   \third{0.829} &     \second{0.108} &              \first{0.9} & 0.00020 \\
NoPE-NeRF*   &           12.44 &           0.390 &              0.770 &                   2607.8 & 0.03342 &           13.06 &           0.460 &              0.710 &                   2608.4 & 0.03761 &           16.42 &           0.390 &              0.680 &                   2469.9 & 0.03112 \\
\midrule
\multicolumn{1}{c|}{} & \multicolumn{5}{|c|}{Church (Tanks \& Temples)} & \multicolumn{5}{|c|}{Courthouse (Tanks \& Temples)} & \multicolumn{5}{|c}{Family (Tanks \& Temples)} \\
\midrule
Method       & PSNR $\uparrow$ & SSIM $\uparrow$ & LPIPS $\downarrow$ & Time (min.) $\downarrow$ & ATE     & PSNR $\uparrow$ & SSIM $\uparrow$ & LPIPS $\downarrow$ & Time (min.) $\downarrow$ & ATE     & PSNR $\uparrow$ & SSIM $\uparrow$ & LPIPS $\downarrow$ & Time (min.) $\downarrow$ & ATE     \\
\midrule
FlowMap      &  \second{28.29} &  \second{0.883} &     \second{0.074} &             \third{22.4} & 0.00061 &           27.51 &   \third{0.911} &      \third{0.055} &             \third{22.2} & 0.00129 &  \second{27.96} &  \second{0.889} &     \second{0.067} &             \third{22.1} & 0.00039 \\
COLMAP       &   \third{27.93} &           0.866 &              0.107 &             \second{6.3} &     N/A &   \third{27.79} &  \second{0.916} &              0.056 &             \second{5.9} &     N/A &           27.13 &   \third{0.878} &              0.092 &             \second{5.0} &     N/A \\
COLMAP (MVS) &   \first{28.71} &   \first{0.890} &      \first{0.068} &                     50.9 &     N/A &   \first{28.56} &   \first{0.926} &      \first{0.044} &                     51.9 &     N/A &   \first{28.40} &   \first{0.897} &      \first{0.062} &                     50.9 &     N/A \\
DROID-SLAM*  &           27.79 &   \third{0.869} &      \third{0.084} &              \first{0.8} & 0.00065 &  \second{27.94} &  \second{0.916} &     \second{0.051} &              \first{0.9} & 0.00034 &   \third{27.78} &           0.873 &      \third{0.081} &              \first{0.8} & 0.00040 \\
NoPE-NeRF*   &           12.91 &           0.400 &              0.700 &                   2575.8 & 0.02752 &           14.92 &           0.510 &              0.590 &                   2599.3 & 0.03462 &           12.87 &           0.470 &              0.700 &                   2597.4 & 0.03232 \\
\midrule
\multicolumn{1}{c|}{} & \multicolumn{5}{|c|}{Francis (Tanks \& Temples)} & \multicolumn{5}{|c|}{Horse (Tanks \& Temples)} & \multicolumn{5}{|c}{Ignatius (Tanks \& Temples)} \\
\midrule
Method       & PSNR $\uparrow$ & SSIM $\uparrow$ & LPIPS $\downarrow$ & Time (min.) $\downarrow$ & ATE     & PSNR $\uparrow$ & SSIM $\uparrow$ & LPIPS $\downarrow$ & Time (min.) $\downarrow$ & ATE     & PSNR $\uparrow$ & SSIM $\uparrow$ & LPIPS $\downarrow$ & Time (min.) $\downarrow$ & ATE     \\
\midrule
FlowMap      &  \second{31.90} &  \second{0.903} &     \second{0.080} &             \third{22.4} & 0.00058 &  \second{28.35} &  \second{0.917} &     \second{0.064} &             \third{22.4} & 0.00054 &   \third{24.54} &   \third{0.773} &     \second{0.131} &             \third{22.4} & 0.00037 \\
COLMAP       &   \third{31.85} &   \third{0.896} &      \third{0.124} &             \second{3.6} &     N/A &           27.34 &           0.903 &              0.097 &             \second{3.4} &     N/A &   \first{24.95} &  \second{0.781} &              0.153 &             \second{5.6} &     N/A \\
COLMAP (MVS) &   \first{32.73} &   \first{0.913} &      \first{0.069} &                     51.1 &     N/A &   \first{28.82} &   \first{0.926} &      \first{0.062} &                     53.2 &     N/A &  \second{24.93} &   \first{0.795} &      \first{0.113} &                     51.2 &     N/A \\
DROID-SLAM*  &           22.23 &           0.753 &              0.275 &              \first{0.9} & 0.00041 &   \third{27.61} &   \third{0.909} &      \third{0.069} &              \first{0.8} & 0.00051 &           24.28 &           0.750 &      \third{0.142} &              \first{0.8} & 0.00025 \\
NoPE-NeRF*   &           17.27 &           0.570 &              0.640 &                    524.9 & 0.02569 &            9.87 &           0.590 &              0.700 &                   2587.4 & 0.04710 &           10.90 &           0.260 &              0.780 &                   2583.2 & 0.04241 \\
\midrule
\multicolumn{1}{c|}{} & \multicolumn{5}{|c|}{M60 (Tanks \& Temples)} & \multicolumn{5}{|c|}{Museum (Tanks \& Temples)} & \multicolumn{5}{|c}{Panther (Tanks \& Temples)} \\
\midrule
Method       & PSNR $\uparrow$ & SSIM $\uparrow$ & LPIPS $\downarrow$ & Time (min.) $\downarrow$ & ATE     & PSNR $\uparrow$ & SSIM $\uparrow$ & LPIPS $\downarrow$ & Time (min.) $\downarrow$ & ATE     & PSNR $\uparrow$ & SSIM $\uparrow$ & LPIPS $\downarrow$ & Time (min.) $\downarrow$ & ATE     \\
\midrule
FlowMap      &   \first{23.23} &   \first{0.805} &      \first{0.190} &             \third{22.4} & 0.00838 &   \third{28.48} &   \third{0.862} &     \second{0.078} &             \third{22.2} & 0.00070 &  \second{27.50} &  \second{0.882} &     \second{0.105} &             \third{22.3} & 0.00112 \\
COLMAP       &   \third{22.04} &  \second{0.803} &      \third{0.219} &             \second{6.2} &     N/A &  \second{28.94} &  \second{0.863} &              0.100 &             \second{5.3} &     N/A &           27.32 &  \second{0.882} &              0.129 &             \second{5.0} &     N/A \\
COLMAP (MVS) &           21.75 &           0.791 &              0.221 &                     51.9 &     N/A &   \first{29.05} &   \first{0.874} &      \first{0.070} &                     50.6 &     N/A &   \first{27.96} &   \first{0.891} &      \first{0.101} &                     52.2 &     N/A \\
DROID-SLAM*  &  \second{22.66} &   \third{0.792} &     \second{0.195} &              \first{0.7} & 0.00667 &           27.74 &           0.833 &      \third{0.096} &              \first{0.8} & 0.00088 &   \third{27.48} &   \third{0.878} &      \third{0.106} &              \first{0.8} & 0.00150 \\
NoPE-NeRF*   &           12.67 &           0.490 &              0.720 &                   2485.1 & 0.04258 &           14.26 &           0.430 &              0.800 &                   2606.9 & 0.03224 &           13.71 &           0.500 &              0.690 &                   2591.0 & 0.03854 \\
\midrule
\multicolumn{1}{c|}{} & \multicolumn{5}{|c|}{Playground (Tanks \& Temples)} & \multicolumn{5}{|c|}{Train (Tanks \& Temples)} & \multicolumn{5}{|c}{Truck (Tanks \& Temples)} \\
\midrule
Method       & PSNR $\uparrow$ & SSIM $\uparrow$ & LPIPS $\downarrow$ & Time (min.) $\downarrow$ & ATE     & PSNR $\uparrow$ & SSIM $\uparrow$ & LPIPS $\downarrow$ & Time (min.) $\downarrow$ & ATE     & PSNR $\uparrow$ & SSIM $\uparrow$ & LPIPS $\downarrow$ & Time (min.) $\downarrow$ & ATE     \\
\midrule
FlowMap      &   \first{24.29} &   \first{0.727} &      \first{0.192} &             \third{22.2} & 0.00096 &   \third{26.22} &   \third{0.870} &      \third{0.077} &             \third{22.2} & 0.00082 &   \third{24.34} &   \third{0.828} &     \second{0.098} &             \third{22.3} & 0.00078 \\
COLMAP       &   \third{22.24} &   \third{0.684} &      \third{0.292} &             \second{7.4} &     N/A &           26.09 &           0.857 &              0.104 &             \second{8.4} &     N/A &  \second{25.57} &  \second{0.848} &      \third{0.104} &             \second{4.9} &     N/A \\
COLMAP (MVS) &  \second{22.92} &  \second{0.693} &     \second{0.230} &                     51.6 &     N/A &   \first{27.43} &   \first{0.888} &      \first{0.063} &                     51.4 &     N/A &   \first{26.39} &   \first{0.864} &      \first{0.080} &                     50.4 &     N/A \\
DROID-SLAM*  &           21.11 &           0.642 &              0.301 &              \first{0.7} & 0.00284 &  \second{26.51} &  \second{0.872} &     \second{0.069} &              \first{0.8} & 0.00088 &           21.48 &           0.739 &              0.208 &              \first{0.8} & 0.00127 \\
NoPE-NeRF*   &           13.53 &           0.360 &              0.770 &                   2613.1 & 0.04120 &           13.18 &           0.440 &              0.670 &                   2614.8 & 0.04052 &           11.71 &           0.410 &              0.740 &                   2603.3 & 0.04583 \\
\midrule
\multicolumn{1}{c|}{} & \multicolumn{5}{|c|}{Bench (CO3D)} & \multicolumn{5}{|c}{Hydrant (CO3D)} \\
\midrule
Method       & PSNR $\uparrow$ & SSIM $\uparrow$ & LPIPS $\downarrow$ & Time (min.) $\downarrow$ & ATE     & PSNR $\uparrow$ & SSIM $\uparrow$ & LPIPS $\downarrow$ & Time (min.) $\downarrow$ & ATE     \\
\midrule
FlowMap      &   \first{33.17} &   \first{0.927} &      \first{0.045} &             \third{22.0} & 0.03094 &           29.05 &           0.865 &              0.083 &             \third{22.1} & 0.00083 \\
COLMAP       &           19.87 &           0.600 &              0.309 &            \second{17.2} &     N/A &  \second{30.46} &  \second{0.900} &     \second{0.070} &             \second{8.0} &     N/A \\
COLMAP (MVS) &   \third{20.00} &   \third{0.616} &      \third{0.292} &                     53.2 &     N/A &   \first{30.70} &   \first{0.908} &      \first{0.057} &                     50.8 &     N/A \\
DROID-SLAM*  &  \second{22.48} &  \second{0.699} &     \second{0.206} &              \first{0.9} & 0.03433 &   \third{29.46} &   \third{0.880} &      \third{0.073} &              \first{0.7} & 0.00024 \\
NoPE-NeRF*   &           13.20 &           0.500 &              0.750 &                   2604.0 & 0.03432 &           16.74 &           0.300 &              0.790 &                   2605.8 & 0.03864 \\
\bottomrule
\end{tabular}

}

\vspace{5pt}
\caption{Results for all individual scenes on all datasets.}
\label{tab:main_comparison_supplemental}
\end{table*}

%% file: figures/depth_vis.tex
\begin{figure}[t!]
    \centering
    \includegraphics[width=\linewidth,]{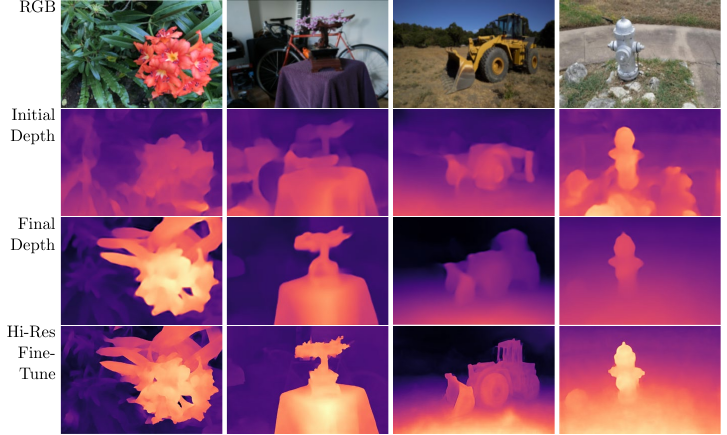}
    \caption{\textbf{Depth Estimates Before and After Optimization.} The depth prediction neural network can either be randomly initialized or pre-trained, though pre-trained depth networks lead to much faster convergence. 
    In the second row, we show the output of the depth prediction neural network after pre-training it on a dataset consisting of CO3D, KITTI, and RealEstate10k. 
    These estimates converge to high-quality depth within only a few hundred FlowMap optimization steps. We see that the quality of the initial, pre-trained depth predictions is not critical to achieve accurate reconstructions.
    Although we estimate geometry at a lower resolution during optimization to manage memory constraints, we can quickly fine-tune at high-resolution for more detailed depth maps if necessary (bottom row). 
    }
    \label{fig:depths}
\end{figure}

%% file: supplement/implementation_details.tex
\section{Implementation Details}

\subsection{Procrustes Solver Details}

Our pose solver is the one introduced in FlowCam \cite{smith2023flowcam}; see \cite{smith2023flowcam} for details. 
The only difference is that instead of selecting 1000 random points for the Procrustes estimation, we fix the points (uniformly spaced throughout the image) when performing per-scene overfitting.
We find that fixing the points used for the pose solver allows the network to better overfit confidence weights and subsequently yields better poses.

\subsection{Intrinsics Solver Details}
For the intrinsics solver, we assume a pinhole camera estimate and discretize a set of 60 candidate focal lengths between .5 and 2 (in resolution-independent units).
We use a softmin on the flow error maps, as discussed in the main paper.
We scale the error maps by a temperature factor of 10 and weight the error maps by the flow confidence weights.
See \cref{fig:focal_and_depth} for illustration.

\subsection{Depth NN (MiDaS) details}
For our depth network, we use the lightweight CNN version of MiDaS \cite{ranftl2020towards}, pretrained with the publicly available weights trained on relative-depth estimation. We optimize the entire network weights during training.

\subsection{Correspondence Weight MLP}
The correspondence weight MLP is a three-layer MLP with ReLU activations and 128 hidden units per layer.
It takes as input two corresponding image features and outputs a per-correspondence weight between 0 and 1 via a sigmoid activation.
Here we use intermediate feature maps from the depth network as the image features.
These weights are used in the weighted Procrustes pose solver.

%% file: supplement/experiment_details.tex
\input{figures/focal_and_depth_est}
\section{Experiment Details}

\subsection{Image Resolution}

To manage computational cost (our current implementation loads the entire video into memory), we compute optical flow and point tracks at a resolution of around 700,000 pixels, then perform FlowMap optimization at 1/16th the resolution.

\subsection{Hyperparameters}

We train for 2000 steps using Adam and use a learning rate of 3e-5.
For the pose-as-variable experiments, we choose Euler angles as the parameterization of the rotation matrix.

\subsection{Pre-Training Details}

Before performing per-scene fine-tuning, we found it useful to learn a large-scale prior for better initialization.
We use the same FlowMap loss formulation but train it on datasets of videos (instead of optimizing on a single scene).
We use videos from CO3D, Real Estate 10K, and KITTI for pretraining.
Note that we only use the raw videos from these datasets (no intrinsics, poses, or sparse geometry).

%% file: figures/focal_and_depth_est.tex
\begin{figure*}[t]
    \centering
    \includegraphics[width=\linewidth,]{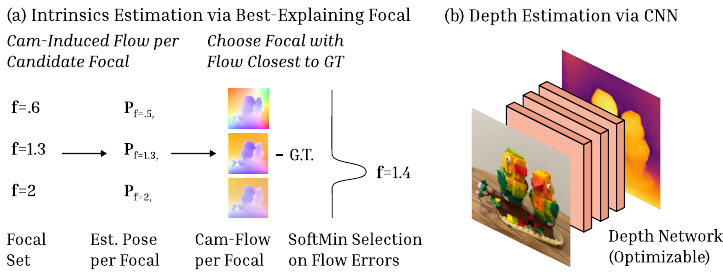}
    \caption{In (a) we illustrate our implicit focal length formulation, which considers a set of candidate focal lengths, assigns each one an error score, and softly selects the focal length with the lowest error. To calculate the error score for a focal length, we use that focal length to estimate a pose, and then compare the resulting pose-induced optical flow to the ground truth optical flow. In (b) we illustrate that we parameterize depth via the output of a monocular depth prediction CNN. }
    \label{fig:focal_and_depth}
\end{figure*}

%% file: supplement/limitations.tex
\section{Limitations}
Wile our method is much faster than MVS COLMAP, it is about 30 percent slower than COLMAP at its highest quality setting (on long sequences, about 20 minutes for our method vs. 14 minutes for COLMAP).
It additionally requires signficantly more GPU memory than COLMAP does.
Our method's pose and intrinsics predictions are less accurate and robust than COLMAP's, as measured by ATE, though after Gaussian Splatting with fine-tuning of camera parameters, we often perform on par with COLMAP.

Our method further depends on correspondences estimated by point tracks and optical flow.
While existing methods for computing point tracks and optical flow are robust, failures sometimes occur, and these failures can affect FlowMap's accuracy if they are significant.
On the other hand, FlowMap will directly improve alongside advancements in these domains.

Finally, our method is constrained to work on frame sequences with significant overlap (i.e., videos) and fails when input sequences contain significant scene motion.
The latter limitation is shared with COLMAP, though we hope that our method may serve as a step towards novel methods that address this shortcoming.

%% file: main.bbl
\begin{thebibliography}{10}
\providecommand{\url}[1]{\texttt{#1}}
\providecommand{\urlprefix}{URL }
\providecommand{\doi}[1]{https://doi.org/#1}

\bibitem{barron2021mipnerf}
Barron, J.T., Mildenhall, B., Tancik, M., Hedman, P., Martin-Brualla, R., Srinivasan, P.P.: Mip-nerf: A multiscale representation for anti-aliasing neural radiance fields. In: Proceedings of the International Conference on Computer Vision (ICCV) (2021)

\bibitem{bian2022nopenerf}
Bian, W., Wang, Z., Li, K., Bian, J., Prisacariu, V.A.: Nope-nerf: Optimising neural radiance field with no pose prior (2023)

\bibitem{bloesch2018codeslam}
Bloesch, M., Czarnowski, J., Clark, R., Leutenegger, S., Davison, A.J.: Codeslam—learning a compact, optimisable representation for dense visual slam. In: Proceedings of the IEEE conference on computer vision and pattern recognition. pp. 2560--2568 (2018)

\bibitem{bowen2022dimensions}
Bowen, R.S., Tucker, R., Zabih, R., Snavely, N.: Dimensions of motion: Monocular prediction through flow subspaces. In: Proceedings of the International Conference on 3D Vision (3DV). pp. 454--464. IEEE (2022)

\bibitem{campos_orb3}
Campos, C., Elvira, R., Rodríguez, J.J.G., M.~Montiel, J.M., D.~Tardós, J.: Orb-slam3: An accurate open-source library for visual, visual–inertial, and multimap slam. Transactions on Robotics (6),  1874--1890 (2021)

\bibitem{chan2023generative}
Chan, E.R., Nagano, K., Chan, M.A., Bergman, A.W., Park, J.J., Levy, A., Aittala, M., De~Mello, S., Karras, T., Wetzstein, G.: Generative novel view synthesis with 3d-aware diffusion models. Proceedings of the International Conference on 3D Vision (3DV)  (2023)

\bibitem{charatan23pixelsplat}
Charatan, D., Li, S., Tagliasacchi, A., Sitzmann, V.: pixelsplat: 3d gaussian splats from image pairs for scalable generalizable 3d reconstruction. In: Proceedings of the IEEE Conference on Computer Vision and Pattern Recognition (CVPR) (2023)

\bibitem{Chen_2023_CVPR}
Chen, Y., Lee, G.H.: Dbarf: Deep bundle-adjusting generalizable neural radiance fields. In: Proceedings of the IEEE/CVF Conference on Computer Vision and Pattern Recognition (CVPR). pp. 24--34 (June 2023)

\bibitem{cheng2023lu}
Cheng, Z., Esteves, C., Jampani, V., Kar, A., Maji, S., Makadia, A.: Lu-nerf: Scene and pose estimation by synchronizing local unposed nerfs. arXiv preprint arXiv:2306.05410  (2023)

\bibitem{chng2022garf}
Chng, S.F., Ramasinghe, S., Sherrah, J., Lucey, S.: Garf: gaussian activated radiance fields for high fidelity reconstruction and pose estimation. arXiv e-prints pp. arXiv--2204 (2022)

\bibitem{choy2020deep}
Choy, C., Dong, W., Koltun, V.: Deep global registration. In: Proc. CVPR (2020)

\bibitem{choy2016universal}
Choy, C.B., Gwak, J., Savarese, S., Chandraker, M.: Universal correspondence network. Advances in neural information processing systems  \textbf{29} (2016)

\bibitem{clark2018learning}
Clark, R., Bloesch, M., Czarnowski, J., Leutenegger, S., Davison, A.J.: Learning to solve nonlinear least squares for monocular stereo. In: Proceedings of the European Conference on Computer Vision (ECCV). pp. 284--299 (2018)

\bibitem{czarnowski2020deepfactors}
Czarnowski, J., Laidlow, T., Clark, R., Davison, A.J.: Deepfactors: Real-time probabilistic dense monocular {SLAM}. Computing Research Repository (CoRR)  (2020)

\bibitem{dsnerf}
Deng, K., Liu, A., Zhu, J.Y., Ramanan, D.: Depth-supervised {NeRF}: Fewer views and faster training for free. In: Proceedings of the IEEE Conference on Computer Vision and Pattern Recognition (CVPR) (June 2022)

\bibitem{detone2018superpoint}
DeTone, D., Malisiewicz, T., Rabinovich, A.: Superpoint: Self-supervised interest point detection and description. In: Proceedings of the IEEE conference on computer vision and pattern recognition workshops. pp. 224--236 (2018)

\bibitem{doersch2023tapir}
Doersch, C., Yang, Y., Vecerik, M., Gokay, D., Gupta, A., Aytar, Y., Carreira, J., Zisserman, A.: Tapir: Tracking any point with per-frame initialization and temporal refinement. arXiv preprint arXiv:2306.08637  (2023)

\bibitem{du2023cross}
Du, Y., Smith, C., Tewari, A., Sitzmann, V.: Learning to render novel views from wide-baseline stereo pairs. In: Proceedings of the IEEE Conference on Computer Vision and Pattern Recognition (CVPR) (2023)

\bibitem{engel2017direct}
Engel, J., Koltun, V., Cremers, D.: Direct sparse odometry. IEEE transactions on pattern analysis and machine intelligence  \textbf{40}(3),  611--625 (2017)

\bibitem{engel2014lsd}
Engel, J., Sch{\"o}ps, T., Cremers, D.: Lsd-slam: Large-scale direct monocular slam. In: Proceedings of the European Conference on Computer Vision (ECCV). pp. 834--849. Springer (2014)

\bibitem{fu2023cbarf}
Fu, H., Yu, X., Li, L., Zhang, L.: Cbarf: Cascaded bundle-adjusting neural radiance fields from imperfect camera poses (2023)

\bibitem{fu2023colmapfree}
Fu, Y., Liu, S., Kulkarni, A., Kautz, J., Efros, A.A., Wang, X.: Colmap-free 3d gaussian splatting  (2023)

\bibitem{fu2022mononerf}
Fu, Y., Misra, I., Wang, X.: Mononerf: Learning generalizable nerfs from monocular videos without camera poses (2023)

\bibitem{10377535}
Gao, Z., Dai, W., Zhang, Y.: Adaptive positional encoding for bundle-adjusting neural radiance fields. In: 2023 IEEE/CVF International Conference on Computer Vision (ICCV). pp. 3261--3271 (2023). \doi{10.1109/ICCV51070.2023.00304}

\bibitem{harley2022particle}
Harley, A.W., Fang, Z., Fragkiadaki, K.: Particle video revisited: {T}racking through occlusions using point trajectories. In: Proceedings of the European Conference on Computer Vision (ECCV) (2022)

\bibitem{jeong2021self}
Jeong, Y., Ahn, S., Choy, C., Anandkumar, A., Cho, M., Park, J.: Self-calibrating neural radiance fields. In: Proceedings of the International Conference on Computer Vision (ICCV). pp. 5846--5854 (2021)

\bibitem{karaev2023cotracker}
Karaev, N., Rocco, I., Graham, B., Neverova, N., Vedaldi, A., Rupprecht, C.: {CoTracker}: It is better to track together (2023)

\bibitem{keetha2023splatam}
Keetha, N., Karhade, J., Jatavallabhula, K.M., Yang, G., Scherer, S., Ramanan, D., Luiten, J.: Splatam: Splat, track \& map 3d gaussians for dense rgb-d slam. arXiv preprint arXiv:2312.02126  (2023)

\bibitem{kerbl20233d}
Kerbl, B., Kopanas, G., Leimk{\"u}hler, T., Drettakis, G.: 3d gaussian splatting for real-time radiance field rendering. ACM Transactions on Graphics (ToG)  \textbf{42}(4),  1--14 (2023)

\bibitem{kingma2014adam}
Kingma, D.P., Ba, J.: Adam: {A} method for stochastic optimization. arXiv:1412.6980  (2014)

\bibitem{Knapitsch2017tanks}
Knapitsch, A., Park, J., Zhou, Q.Y., Koltun, V.: Tanks and temples: Benchmarking large-scale scene reconstruction. ACM Transactions on Graphics (TOG)  \textbf{36}(4) (2017)

\bibitem{kopf2021robust}
Kopf, J., Rong, X., Huang, J.B.: Robust consistent video depth estimation. In: Proceedings of the IEEE Conference on Computer Vision and Pattern Recognition (CVPR). pp. 1611--1621 (2021)

\bibitem{lai2021video}
Lai, Z., Liu, S., Efros, A.A., Wang, X.: Video autoencoder: self-supervised disentanglement of static 3d structure and motion. In: Proceedings of the International Conference on Computer Vision (ICCV). pp. 9730--9740 (2021)

\bibitem{li2023neuralangelo}
Li, Z., M\"uller, T., Evans, A., Taylor, R.H., Unberath, M., Liu, M.Y., Lin, C.H.: Neuralangelo: High-fidelity neural surface reconstruction. In: IEEE Conference on Computer Vision and Pattern Recognition ({CVPR}) (2023)

\bibitem{lin2021barf}
Lin, C.H., Ma, W.C., Torralba, A., Lucey, S.: Barf: Bundle-adjusting neural radiance fields. In: Proceedings of the International Conference on Computer Vision (ICCV). pp. 5741--5751 (2021)

\bibitem{lindenberger2023lightglue}
Lindenberger, P., Sarlin, P.E., Pollefeys, M.: {LightGlue: Local Feature Matching at Light Speed}. In: Proceedings of the International Conference on Computer Vision (ICCV) (2023)

\bibitem{liu2019neural}
Liu, C., Gu, J., Kim, K., Narasimhan, S.G., Kautz, J.: Neural rgb (r) d sensing: Depth and uncertainty from a video camera. In: Proceedings of the IEEE Conference on Computer Vision and Pattern Recognition (CVPR). pp. 10986--10995 (2019)

\bibitem{liu2023robust}
Liu, Y.L., Gao, C., Meuleman, A., Tseng, H.Y., Saraf, A., Kim, C., Chuang, Y.Y., Kopf, J., Huang, J.B.: Robust dynamic radiance fields. In: Proceedings of the IEEE/CVF Conference on Computer Vision and Pattern Recognition. pp. 13--23 (2023)

\bibitem{sift}
Lowe, D.: Object recognition from local scale-invariant features. In: Proceedings of the Seventh IEEE International Conference on Computer Vision. pp. 1150--1157 vol.2 (1999)

\bibitem{luo2018geodesc}
Luo, Z., Shen, T., Zhou, L., Zhu, S., Zhang, R., Yao, Y., Fang, T., Quan, L.: Geodesc: Learning local descriptors by integrating geometry constraints. In: Proceedings of the European Conference on Computer Vision (ECCV). pp. 168--183 (2018)

\bibitem{Matsuki:Murai:etal:CVPR2024}
Matsuki, H., Murai, R., Kelly, P.H.J., Davison, A.J.: {G}aussian {S}platting {SLAM}. In: Proceedings of the IEEE Conference on Computer Vision and Pattern Recognition (CVPR) (2024)

\bibitem{meuleman2023localrf}
Meuleman, A., Liu, Y.L., Gao, C., Huang, J.B., Kim, C., Kim, M.H., Kopf, J.: Progressively optimized local radiance fields for robust view synthesis. In: CVPR (2023)

\bibitem{mildenhall2019local}
Mildenhall, B., Srinivasan, P.P., Ortiz-Cayon, R., Kalantari, N.K., Ramamoorthi, R., Ng, R., Kar, A.: Local light field fusion: Practical view synthesis with prescriptive sampling guidelines. ACM Transactions on Graphics (TOG)  \textbf{38}(4),  1--14 (2019)

\bibitem{mildenhall2020nerf}
Mildenhall, B., Srinivasan, P.P., Tancik, M., Barron, J.T., Ramamoorthi, R., Ng, R.: {NeRF}: {R}epresenting scenes as neural radiance fields for view synthesis. In: Proceedings of the European Conference on Computer Vision (ECCV). pp. 405--421 (2020)

\bibitem{mishchuk2017working}
Mishchuk, A., Mishkin, D., Radenovic, F., Matas, J.: Working hard to know your neighbor's margins: Local descriptor learning loss. Advances in Neural Information Processing Systems (NeurIPS)  \textbf{30} (2017)

\bibitem{mueller2022instant}
M\"uller, T., Evans, A., Schied, C., Keller, A.: Instant neural graphics primitives with a multiresolution hash encoding. ACM Transactions on Graphics (TOG)  \textbf{41}(4),  102:1--102:15 (2022)

\bibitem{mur2015orb}
Mur-Artal, R., Montiel, J.M.M., Tardos, J.D.: Orb-slam: a versatile and accurate monocular slam system. Transactions on Robotics (5),  1147--1163 (2015)

\bibitem{mur2017orb}
Mur-Artal, R., Tard{\'o}s, J.D.: Orb-slam2: An open-source slam system for monocular, stereo, and rgb-d cameras. Transactions on Robotics  \textbf{33}(5),  1255--1262 (2017)

\bibitem{niemeyer2020differentiable}
Niemeyer, M., Mescheder, L., Oechsle, M., Geiger, A.: Differentiable volumetric rendering: Learning implicit 3d representations without 3d supervision. In: Proceedings of the IEEE Conference on Computer Vision and Pattern Recognition (CVPR). pp. 3504--3515 (2020)

\bibitem{ono2018lf}
Ono, Y., Trulls, E., Fua, P., Yi, K.M.: Lf-net: Learning local features from images. Advances in Neural Information Processing Systems (NeurIPS)  \textbf{31} (2018)

\bibitem{park2023camp}
Park, K., Henzler, P., Mildenhall, B., Barron, J.T., Martin-Brualla, R.: Camp: Camera preconditioning for neural radiance fields. ACM Transactions on Graphics (TOG)  \textbf{42}(6),  1--11 (2023)

\bibitem{ranftl2020towards}
Ranftl, R., Lasinger, K., Hafner, D., Schindler, K., Koltun, V.: Towards robust monocular depth estimation: {M}ixing datasets for zero-shot cross-dataset transfer. IEEE Transactions on Pattern Analysis and Machine Intelligence  (2020)

\bibitem{reizenstein2021common}
Reizenstein, J., Shapovalov, R., Henzler, P., Sbordone, L., Labatut, P., Novotny, D.: Common objects in 3d: Large-scale learning and evaluation of real-life 3d category reconstruction. In: Proceedings of the International Conference on Computer Vision (ICCV). pp. 10901--10911 (2021)

\bibitem{rosinol2020kimera}
Rosinol, A., Abate, M., Chang, Y., Carlone, L.: Kimera: an open-source library for real-time metric-semantic localization and mapping. In: Proceedings of the IEEE International Conference on Robotics and Automation (ICRA). pp. 1689--1696. IEEE (2020)

\bibitem{sarlin2020superglue}
Sarlin, P.E., DeTone, D., Malisiewicz, T., Rabinovich, A.: Superglue: Learning feature matching with graph neural networks. In: Proceedings of the IEEE Conference on Computer Vision and Pattern Recognition (CVPR). pp. 4938--4947 (2020)

\bibitem{sarlin20superglue}
Sarlin, P.E., DeTone, D., Malisiewicz, T., Rabinovich, A.: {SuperGlue}: {L}earning feature matching with graph neural networks. In: Proceedings of the IEEE Conference on Computer Vision and Pattern Recognition (CVPR) (2020)

\bibitem{sarlin2021pixloc}
Sarlin, P.E., Unagar, A., Larsson, M., Germain, H., Toft, C., Larsson, V., Pollefeys, M., Lepetit, V., Hammarstrand, L., Kahl, F., Sattler, T.: {Back to the Feature: Learning Robust Camera Localization from Pixels to Pose}. In: Proceedings of the IEEE Conference on Computer Vision and Pattern Recognition (CVPR) (2021)

\bibitem{schonberger2016structure}
Schonberger, J.L., Frahm, J.M.: Structure-from-motion revisited. In: Proceedings of the IEEE Conference on Computer Vision and Pattern Recognition (CVPR). pp. 4104--4113 (2016)

\bibitem{sitzmann2019deepvoxels}
Sitzmann, V., Thies, J., Heide, F., Nie{\ss}ner, M., Wetzstein, G., Zollh{\"o}fer, M.: Deepvoxels: Learning persistent 3d feature embeddings. In: Proceedings of the IEEE Conference on Computer Vision and Pattern Recognition (CVPR) (2019)

\bibitem{sitzmann2019scene}
Sitzmann, V., Zollh{\"o}fer, M., Wetzstein, G.: Scene representation networks: Continuous 3d-structure-aware neural scene representations. Advances in Neural Information Processing Systems (NeurIPS)  (2019)

\bibitem{smith2023flowcam}
Smith, C., Du, Y., Tewari, A., Sitzmann, V.: Flowcam: Training generalizable 3d radiance fields without camera poses via pixel-aligned scene flow. Advances in Neural Information Processing Systems (NeurIPS)  (2023)

\bibitem{suhail2022generalizable}
Suhail, M., Esteves, C., Sigal, L., Makadia, A.: Generalizable patch-based neural rendering. In: Proceedings of the European Conference on Computer Vision (ECCV). Springer (2022)

\bibitem{nerfstudio}
Tancik, M., Weber, E., Ng, E., Li, R., Yi, B., Kerr, J., Wang, T., Kristoffersen, A., Austin, J., Salahi, K., Ahuja, A., McAllister, D., Kanazawa, A.: Nerfstudio: A modular framework for neural radiance field development. In: ACM Transactions on Graphics (TOG) (2023)

\bibitem{tang2018ba}
Tang, C., Tan, P.: {BA-Net}: {D}ense bundle adjustment network. arXiv preprint arXiv:1806.04807  (2018)

\bibitem{teed2018deepv2d}
Teed, Z., Deng, J.: Deepv2d: Video to depth with differentiable structure from motion. arXiv preprint arXiv:1812.04605  (2018)

\bibitem{raft}
Teed, Z., Deng, J.: {RAFT}: {R}ecurrent all-pairs field transforms for optical flow. In: Proceedings of the European Conference on Computer Vision (ECCV) (2020)

\bibitem{teed2021droid}
Teed, Z., Deng, J.: Droid-slam: Deep visual slam for monocular, stereo, and rgb-d cameras. Advances in Neural Information Processing Systems (NeurIPS)  \textbf{34} (2021)

\bibitem{tewari2023diffusion}
Tewari, A., Yin, T., Cazenavette, G., Rezchikov, S., Tenenbaum, J.B., Durand, F., Freeman, W.T., Sitzmann, V.: Diffusion with forward models: Solving stochastic inverse problems without direct supervision. Advances in Neural Information Processing Systems (NeurIPS)  (2023)

\bibitem{ummenhofer2017demon}
Ummenhofer, B., Zhou, H., Uhrig, J., Mayer, N., Ilg, E., Dosovitskiy, A., Brox, T.: Demon: Depth and motion network for learning monocular stereo. In: Proceedings of the IEEE Conference on Computer Vision and Pattern Recognition (CVPR). pp. 5038--5047 (2017)

\bibitem{wang2023visual}
Wang, J., Karaev, N., Rupprecht, C., Novotny, D.: Visual geometry grounded deep structure from motion. arXiv preprint arXiv:2312.04563  (2023)

\bibitem{wang2021ibrnet}
Wang, Q., Wang, Z., Genova, K., Srinivasan, P., Zhou, H., Barron, J.T., Martin-Brualla, R., Snavely, N., Funkhouser, T.: Ibrnet: Learning multi-view image-based rendering. In: Proceedings of the IEEE Conference on Computer Vision and Pattern Recognition (CVPR) (2021)

\bibitem{wang2021tartanvo}
Wang, W., Hu, Y., Scherer, S.: Tartanvo: A generalizable learning-based vo. In: Conference on Robot Learning. pp. 1761--1772. PMLR (2021)

\bibitem{nerf--}
Wang, Z., Wu, S., Xie, W., Chen, M., Prisacariu, V.A.: Ne{RF}$--$: Neural radiance fields without known camera parameters. arXiv:2102.07064  (2021)

\bibitem{wang2021nerf}
Wang, Z., Wu, S., Xie, W., Chen, M., Prisacariu, V.A.: Nerf--: Neural radiance fields without known camera parameters. arXiv preprint arXiv:2102.07064  (2021)

\bibitem{wewer24latentsplat}
Wewer, C., Raj, K., Ilg, E., Schiele, B., Lenssen, J.E.: latentsplat: Autoencoding variational gaussians for fast generalizable 3d reconstruction. In: arXiv (2024)

\bibitem{wu2023scanerf}
Wu, X., Xu, J., Zhang, X., Bao, H., Huang, Q., Shen, Y., Tompkin, J., Xu, W.: Scanerf: Scalable bundle-adjusting neural radiance fields for large-scale scene rendering. ACM Transactions on Graphics (TOG)  (2023)

\bibitem{xia2022sinerf}
Xia, Y., Tang, H., Timofte, R., Van~Gool, L.: Sinerf: Sinusoidal neural radiance fields for joint pose estimation and scene reconstruction. arXiv preprint arXiv:2210.04553  (2022)

\bibitem{xu2022gmflow}
Xu, H., Zhang, J., Cai, J., Rezatofighi, H., Tao, D.: Gmflow: Learning optical flow via global matching. In: Proceedings of the IEEE/CVF Conference on Computer Vision and Pattern Recognition. pp. 8121--8130 (2022)

\bibitem{yan2023cf}
Yan, Q., Wang, Q., Zhao, K., Chen, J., Li, B., Chu, X., Deng, F.: Cf-nerf: Camera parameter free neural radiance fields with incremental learning. arXiv preprint arXiv:2312.08760  (2023)

\bibitem{yen2021inerf}
Yen-Chen, L., Florence, P., Barron, J.T., Rodriguez, A., Isola, P., Lin, T.Y.: inerf: Inverting neural radiance fields for pose estimation. In: 2021 IEEE/RSJ International Conference on Intelligent Robots and Systems (IROS). pp. 1323--1330. IEEE (2021)

\bibitem{pixelnerf}
Yu, A., Ye, V., Tancik, M., Kanazawa, A.: {pixelNeRF}: {N}eural radiance fields from one or few images. In: Proceedings of the IEEE Conference on Computer Vision and Pattern Recognition (CVPR) (2021)

\bibitem{yugay2023gaussian}
Yugay, V., Li, Y., Gevers, T., Oswald, M.R.: Gaussian-slam: Photo-realistic dense slam with gaussian splatting. arXiv preprint arXiv:2312.10070  (2023)

\bibitem{zhang2022casual}
Zhang, Z., Cole, F., Li, Z., Rubinstein, M., Snavely, N., Freeman, W.T.: Structure and motion from casual videos. In: Proceedings of the European Conference on Computer Vision (ECCV). pp. 20--37. Springer (2022)

\bibitem{zhao2022particlesfm}
Zhao, W., Liu, S., Guo, H., Wang, W., Liu, Y.J.: Particlesfm: Exploiting dense point trajectories for localizing moving cameras in the wild. In: Proceedings of the European Conference on Computer Vision (ECCV). pp. 523--542. Springer (2022)

\bibitem{zhou2018deeptam}
Zhou, H., Ummenhofer, B., Brox, T.: Deeptam: Deep tracking and mapping. In: Proceedings of the European Conference on Computer Vision (ECCV). pp. 822--838 (2018)

\end{thebibliography}
